\documentclass[lettersize,journal]{IEEEtran}
\usepackage{amsmath,amsfonts}
\usepackage{algorithmic}
\usepackage{algorithm}
\usepackage{array}
\usepackage{textcomp}
\usepackage{stfloats}
\usepackage{url}
\usepackage{verbatim}
\usepackage{graphicx}
\usepackage{cite}

\usepackage{amsmath,amsfonts,bm}

\def\eqref#1{equation~\ref{#1}}

\def\1{\bm{1}}

\def\vx{{\bm{x}}}
\def\vy{{\bm{y}}}

\DeclareMathAlphabet{\mathsfit}{\encodingdefault}{\sfdefault}{m}{sl}
\SetMathAlphabet{\mathsfit}{bold}{\encodingdefault}{\sfdefault}{bx}{n}

\usepackage{hyperref}
\usepackage[numbers]{natbib}
\usepackage{url}
\usepackage{subfigure}
\usepackage{hyperref}
\usepackage{multirow}
\usepackage{booktabs}
\usepackage{makecell}
\usepackage{rotating}
\usepackage{pifont}
\usepackage{multicol}
\usepackage{color}
\usepackage{colortbl}
\usepackage{xcolor}
\usepackage{soul}
\usepackage{wrapfig}
\newcommand*\colourcheck[1]{%
  \expandafter\newcommand\csname #1check\endcsname{\textcolor{#1}{\ding{52}}}%
}
\colourcheck{green}

\newcommand*\colourcross[1]{%
  \expandafter\newcommand\csname #1cross\endcsname{\textcolor{#1}{\ding{56}}}%
}
\colourcross{red}

\newcommand{\red}{\textcolor{red}}
\newcommand{\blue}{\textcolor{blue}}
\newcommand{\scr}{\scriptsize}
\newcommand{\fs}{\tiny}

\begin{document}

\title{Towards Mitigating Architecture Overfitting on Distilled Datasets}
\author{\IEEEauthorblockN{Xuyang Zhong, Chen Liu\IEEEauthorrefmark{1}}
        
        \IEEEauthorblockA{Department of Computer Science,
City University of Hong Kong, Hong Kong SAR, China\\
xuyang.zhong@my.cityu.edu.hk, chen.liu@cityu.edu.hk}
\thanks{\IEEEauthorrefmark{1} denotes the correspondence author.}
}



\maketitle

\begin{abstract}
    Dataset distillation methods have demonstrated remarkable performance for neural networks trained with very limited training data. However, a significant challenge arises in the form of \textit{architecture overfitting}: the distilled training dataset synthesized by a specific network architecture (i.e., training network) generates poor performance when trained by other network architectures (i.e., test networks), especially when the test networks have a larger capacity than the training network. This paper introduces a series of approaches to mitigate this issue. Among them, DropPath renders the large model to be an implicit ensemble of its sub-networks, and knowledge distillation ensures each sub-network acts similarly to the small but well-performing teacher network. These methods, characterized by their smoothing effects, significantly mitigate architecture overfitting. We conduct extensive experiments to demonstrate the effectiveness and generality of our methods. Particularly, across various scenarios involving different tasks and different sizes of distilled data, our approaches significantly mitigate architecture overfitting. Furthermore, our approaches achieve comparable or even superior performance when the test network is larger than the training network. Codes are available at \href{https://github.com/CityU-MLO/mitigate_architecture_overfitting}{CityU-MLO/mitigate\_architecture\_overfitting}.
\end{abstract}


\section{Introduction}

Deep learning has achieved tremendous success in various applications \citep{rombach2022highresolution, JumperEvansPritzel2021}, but training a powerful deep neural network requires massive training data \citep{dosovitskiy2020image, brown2020language}.
To accelerate training, one possible way is to construct a new but smaller training set that preserves most of the information of the original larger set.
In this regard, we can use \textit{coreset} \citep{coleman2019selection, hwang2020data} to sample a subset of the original training set or \textit{dataset distillation} \citep{wang2018dataset, zhao2020dataset} to synthesize a small training set. Compared with coreset, dataset distillation is demonstrated to achieve much better performance when the amount of data is extremely small \citep{hwang2020data, zhao2023dataset}. Furthermore, dataset distillation is shown to benefit various applications, such as continual learning \citep{zhao2020dataset, rosasco2022distilled, zhao2021dataset, zhao2023dataset}, neural architecture search \citep{zhao2020dataset, zhao2021dataset}, and privacy preservation \citep{li2020soft, goetz2020federated}. Therefore, in this work, we focus on dataset distillation to compress the training set.

In the dataset distillation framework, the small training set, which is also called the \textit{distilled dataset}, is learned by using a neural network, which we call \textit{training network}, to extract the most important information from the original training set. Existing data distillation methods are based on various techniques, including meta-learning~\citep{wang2018dataset, bohdal2020flexible, sucholutsky2021soft, nguyen2021kip, nguyen2021dataset, zhou2022dataset} and data matching~\citep{zhao2021dataset, lee2022dataset, zhao2020dataset,cazenavette2022dataset, cui2022scaling, wang2022cafe, zhao2023dataset}.
These methods are then evaluated by the test accuracy of another neural network, which we call \textit{test network}, trained on the distilled dataset.
In summary, in the context of dataset distillation, the training network serves as the model utilized for constructing the distilled dataset, while the test network is employed to showcase the performance achievable through the distilled dataset.

Despite efficiency, dataset distillation methods generally suffer from \textit{architecture overfitting} \citep{zhao2021dataset, zhao2023dataset, nguyen2021dataset, zhou2022dataset, cazenavette2022dataset}.
That is, the performance of the test network trained on the distilled dataset degrades significantly when it has a different network architecture from the training network.
Moreover, the performance deteriorates further when there is a larger difference between the training and test networks in terms of depth and topological structure.
Specifically, due to high computational complexity and optimization challenges in dataset distillation, the training networks are usually shallow networks, such as 3-layer convolutional neural networks (CNN) as used in \citet{cazenavette2022dataset, zhou2022dataset}.
However, such shallow networks are rarely employed in practical applications due to their limited representation power. Consequently, we posit that the architecture overfitting seriously undermines the practicality of distilled datasets in real-world scenarios.

Our analysis in this work indicates that the performance gap between different network architectures is larger in the case of training on the distilled dataset than in the case of training on the subset of the original training set.
In addition, compared with methods compressing the training set by subset selection, dataset distillation achieves better performance when using the same amount of training instances and is thus more popular in downstream applications \citep{zhao2021dataset, zhao2023dataset, goetz2020federated}.
Therefore, we mainly focus on dataset distillation, in which the effectiveness of the proposed method can be better revealed.

\begin{figure*}[!t]
\small
    \centering
    \subfigure[When we use FRePo \citep{zhou2022dataset} to construct the distilled dataset.]{\includegraphics[width=0.4\textwidth]{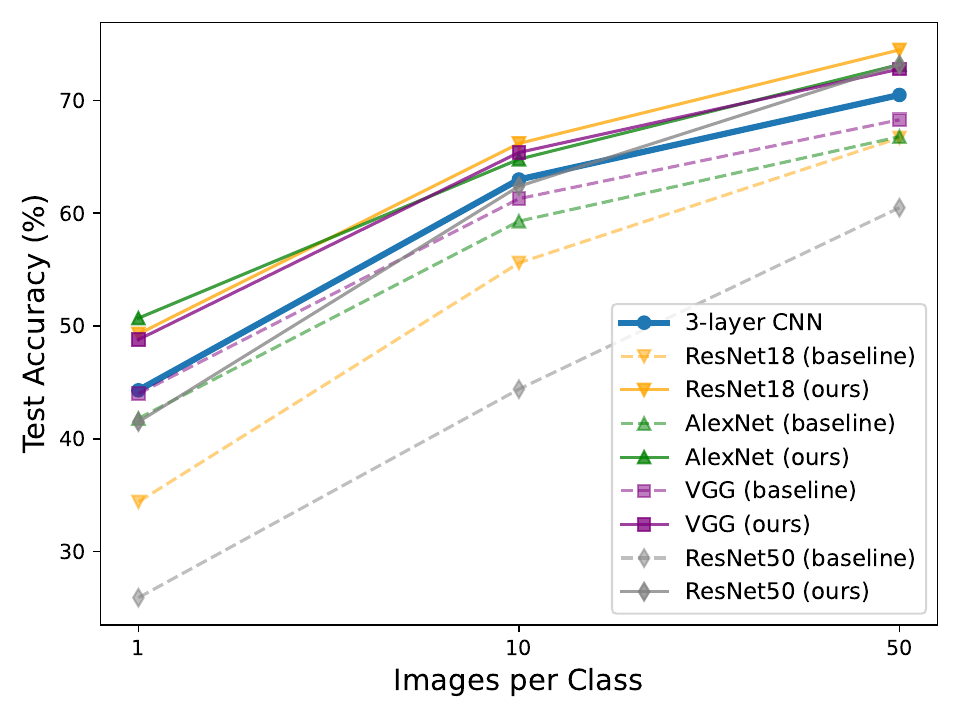}}
    ~~~~~~~~
    \subfigure[When we use MTT \citep{cazenavette2022dataset} to construct the distilled dataset.]
    {\includegraphics[width=0.4\textwidth]{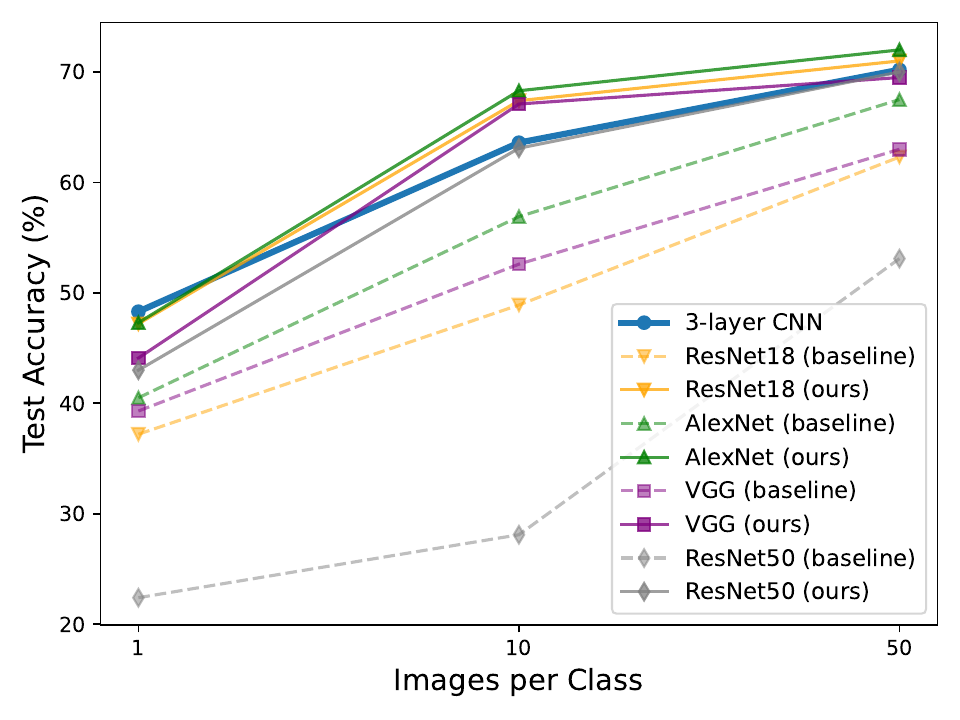}}
    \caption{\small Effectiveness of our method on different architectures, different dataset distillation methods, and different images per class (IPCs) on CIFAR10. 
    We use a 3-layer CNN as the training network, so it performs the best among various architectures in baselines (dashed lines).
    Our methods (solid lines) can significantly narrow down the performance gap between the 3-layer CNN and other architectures.
    }\label{fig:1}
\end{figure*}

In this work, we demonstrate that the architecture overfitting issue on distilled datasets can be mitigated by a better architecture design and training scheme of test networks on the distilled dataset.
Firstly, we combine DropPath with knowledge distillation from a small teacher network.
Specifically, DropPath renders the large model to be an implicit ensemble of its sub-networks, and knowledge distillation ensures each sub-network acts similarly to the small but well-performing teacher network. As a result, the large models could outperform small teacher models on distilled datasets. Additionally, we propose a series of approaches, including three-phase DropPath keep rate, improved shortcut connection, periodical learning rates, a better optimizer and a stronger augmentation scheme, to further boost the performance. These methods share a common characteristic of smoothing the optimization problem from different aspects.
Notably, our proposed methods are also generic: we conduct comprehensive experiments on different network architectures, different numbers of instances per class (IPC), different dataset distillation methods and different datasets to demonstrate the effectiveness of our methods.
Figure~\ref{fig:1} above demonstrates the performance of our proposed methods in various scenarios. It is clear that our methods greatly mitigate architecture overfitting and make large networks trained on distilled datasets achieve better performance in most cases. As a result, the utility and transferability of distilled datasets in practice is markedly enhanced even without modifying the dataset distillation algorithm.
In addition to dataset distillation, our methods can also improve the performance of training on a small real dataset, including those constructed by coresets.
Although some tasks, like synthetic-to-real generalization \citep{chen2021contrastive} and few-shot learning \citep{parnami2022learning}, are also classical problems, customizing our method for them is out of the scope of this work, because we focus on training large networks on small datasets from scratch. We leave these tasks as future works.

We summarize the contributions of this paper as follows: 
\begin{enumerate}
    \item We propose a series of approaches to mitigate architecture overfitting on distilled datasets. Among them, DropPath renders the large model to be an implicit ensemble of its sub-networks, and knowledge distillation ensures each sub-network acts similarly to the small teacher network. These methods share a common characteristic of smoothing the optimization problem. Our proposed methods are plug-and-play and applicable to different model architectures and training schemes.

    \item We conduct extensive experiments to demonstrate that our method significantly mitigates architecture overfitting across different network architectures, different dataset distillation approaches, different numbers of instances per class (IPC), and different datasets.

    \item Moreover, our method generally improves the performance of deep networks trained on limited real data. As a result, large networks outperform small networks on various amounts of training data, even when there are only $100$ training samples.
\end{enumerate}

\section{Related Works}
\textbf{Dataset Distillation:} The goal of dataset distillation is to learn a smaller set of training samples called \textit{distilled dataset} that preserves essential information of the original large dataset so that models trained on this small dataset have similar performance to those trained on the original large dataset.

Existing dataset distillation approaches are based on either meta-learning or data matching \citep{lei2023comprehensive}.
The former category includes backpropagation through time (BPTT) approach \citep{wang2018dataset, bohdal2020flexible, sucholutsky2021soft} and kernel ridge regression (KRR) approach \citep{nguyen2021kip, nguyen2021dataset, zhou2022dataset, loo2022efficient, loo2023dataset}; the latter category includes gradient matching \citep{zhao2021dataset, lee2022dataset}, trajectory matching \citep{cazenavette2022dataset, du2022minimizing, cui2022scaling, guotowards, du2023seqmatch, lee2024selmatch, liu2024att, yang2024nsd}, and distribution matching \citep{wang2022cafe, zhao2023dataset, Sajedi_2023_ICCV, zhao2023idm, zhang2024m3d, deng2024iid, zhang2024dance, li2024dsdm}.
In addition, some works \citep{zhang2023accelerating, liu2023dream, he2023yoco, shang2023mim4dd, chen2024vodka, yunzhen2024embarassingly, he2024mdc, loo2024d3s, xu2024distill} leverage better optimization schemes to improve the performance of dataset distillation.
However, these methods are shown to suffer from severe \textit{architecture overfitting}: the significant performance degradation when the architecture of the training network and the test network are different.
Recently, some factorization methods \citep{kim2022dataset, deng2022remember, liu2022dataset, lee2022datasetfactor, wei2023sparse, shin2023fred, zheng2024hmn}, which learn synthetic datasets by optimizing their factorized features and corresponding decoders, greatly improve the cross-architecture transferability.
However, the instance per class (IPC), which indicates the size of the distilled dataset, used in these methods is much larger than that of meta-learning and data matching approaches, which greatly cancels out the advantages of dataset distillation.
To better fit the motivation of dataset distillation, we only consider small IPCs ($1$, $10$ and $50$) in this work, so the factorization methods are not included for comparison.

\textbf{Model Ensemble:} Model ensemble aims to integrate multiple models to improve the generalization performance.
Popular ensemble methods for classification models include bagging \citep{breiman1996bagging}, AdaBoost \citep{hastie2009multi}, random forest \citep{breiman2001random}, random subspace \citep{ho1995random}, and gradient boosting \citep{friedman2002stochastic}.
However, these methods require training several models and thus are computationally expensive.
By contrast, DropOut \citep{srivastava2014dropout} trains the model only once but stochastically masks its intermediate feature maps during training.
At each training iteration with DropOut, only part of the model parameters are updated, which forms a sub-network of the model.
In this regard, DropOut enables implicit model ensembles of different sub-networks to improve the generalization performance.
Similar to DropOut, DropPath \citep{larsson2016fractalnet} also implicitly ensembles sub-networks but it blocks a whole layer rather than masking some feature maps.
Therefore, it is applicable to network architectures with multiple branches, such as ResNet \citep{he2016deep}, otherwise, the model output will be zero if a layer of a single branch network is dropped.
By contrast, we propose a DropPath variant in this work which is generic, applicable to single-branch networks and effective in mitigating architecture overfitting.

\textbf{Knowledge Distillation:} 
Knowledge distillation \citep{hinton2015distilling} aims to compress a well-trained large model (i.e., teacher model) into a smaller and more efficient model (i.e., student model) with comparable performance.
The standard knowledge distillation \citep{hinton2015distilling} is also known as offline distillation since the teacher model is fixed when training the student model.
Online distillation \citep{zhang2018deep, chen2020online} is proposed to further improve the performance of the student model, especially when a large-capacity high-performance teacher model is not available.
In online distillation, both the teacher model and the student model are updated simultaneously.
In most cases, knowledge distillation methods use large models as the teachers and small models as the students, which is based on the fact that larger models typically have better performance.
However, in the context of dataset distillation, a smaller test network with the same architecture as the training network can achieve a better performance than a larger one on the distilled dataset, so we use the small model as the teacher and the large model as the student in this work. In this way, the performance of large models trained on distilled datasets can be boosted significantly.

We show in the following sections that combining DropPath and knowledge distillation, architecture overfitting on distilled datasets can be almost overcome.
\section{Methods}
In this section, we introduce the approaches that are effective in mitigating architecture overfitting on distilled datasets.
Our methods are motivated by the intuition that the large model can act as an implicit ensemble of small models \citep{srivastava2014dropout, larsson2016fractalnet}.
First, we propose a DropPath variant, which implicitly ensemble sub-networks of models and is different from vanilla DropPath \citep{larsson2016fractalnet} in that the proposed DropPath variant is also applicable to single-branch architectures.
Correspondingly, we optimize the shortcut connections of ResNet-like architecture to accommodate DropPath better.
Second, we use knowledge distillation \citep{hinton2015distilling} as a form of regularization to ensure each sub-network induced by DropPath acts similarly to the teacher network. In contrast to traditional knowledge distillation approaches \citep{hinton2015distilling, zhang2018deep}, the teacher model is smaller than the student model in our cases.
Finally, we adopt a periodical learning rate scheduler, a gradient symbol-based optimizer \citep{chen2023symbolic}, and a stronger data augmentation scheme to improve the performance further. 

\subsection{DropPath with Three-Phase Keep Rate} \label{sec:droppath}
\begin{figure*}[!ht]
\small
    \centering
    \subfigure[Multi-branch]{\includegraphics[width=0.24\textwidth, height=0.3\textwidth]{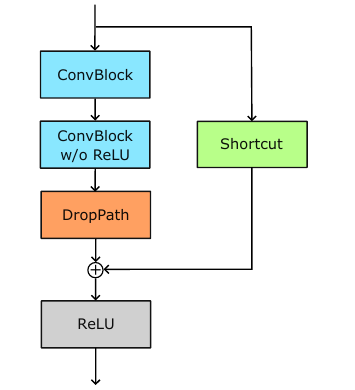}} 
    \subfigure[Single-branch]{\includegraphics[width=0.24\textwidth, height=0.3\textwidth]{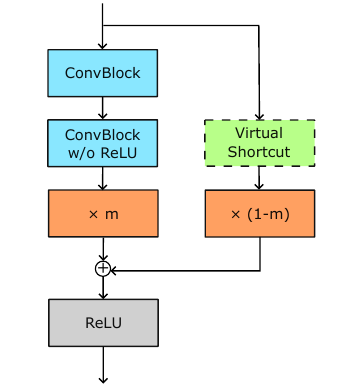}} 
    \subfigure[Original shortcut]{\includegraphics[width=0.24\textwidth, height=0.3\textwidth]{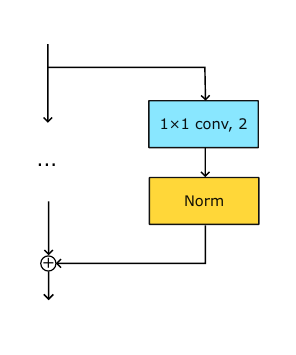}} 
    \subfigure[Improved shortcut]{\includegraphics[width=0.24\textwidth, height=0.3\textwidth]{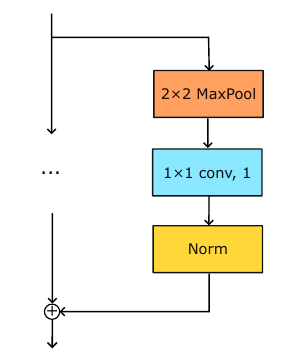}}
    \caption{\small \textbf{(a)} The DropPath used for multi-branch residual blocks during training, it does not block the shortcut path.
    \textbf{(b)} The DropPath used for single-branch networks during training. Here, $m=\mathtt{Bernoulli}(p) \in \{0,1\}$, $p\in[0,1]$ denotes the keep rate. Only when the main path is pruned ($m = 0$), the virtual shortcut is activated, and vice versa. DropPath is always deactivated, i.e., $p=1$, during inference.
    \textbf{(c)} The original architecture of a shortcut connection to downsample feature maps, which consists of a $1\times1$ convolution layer with the stride of $2$ and a normalization layer. \textbf{(d)} The improved architecture of a shortcut connection to downsample feature maps, which is a sequence of a $2 \times 2$ max pooling layer, a $1 \times 1$ convolution layer with the stride of $1$ and a normalization layer.}
    \label{fig:dp}
\end{figure*}

Similar to DropOut \citep{srivastava2014dropout}, DropPath \citep{larsson2016fractalnet}, a.k.a., stochastic depth, was proposed to improve generalization.
While DropOut masks some entries of feature maps, DropPath randomly prunes the entire branch in a multi-branch architecture.
To obtain a deterministic model for evaluation, DropPath is deactivated during inference.
To ensure the expectation of the feature maps to be consistent for training and inference, we scale the output of feature maps after DropPath during training.
Mathematically, DropPath works as follows:
\begin{equation} \label{eq:dp}
    \mathtt{DropPath}(\vx) = \frac{m}{p}\cdot \vx,\quad m=\mathtt{Bernoulli}(p).
\end{equation}
where $p \in [0, 1]$ denotes the keep rate, $m = \mathtt{Bernoulli}(p) \in \{0,1\}$ outputs 1 with probability $p$ and 0 with probability $1-p$. 
The scaling factor $1 / p$ is used to ensure the expectation of the feature maps remains unchanged after DropPath. The detailed derivation is in Appendix \ref{sec:ap_scale_factor}.
Figure \ref{fig:dp} (a) illustrates how DropPath is integrated into networks. 
It effectively decreases the model complexity during training and can force the model to learn more generalizable representations using fewer layers.
Same as DropOut, any network trained with DropPath can be regarded as an ensemble of its subnetworks \citep{NIPS2012_c399862d}, which has been proven to improve generalization \citep{breiman1996bagging, hastie2009multi, breiman2001random, ho1995random, friedman2002stochastic}.
Note that, DropOut masks part of the feature maps and effectively decreases the network width; by contrast, DropPath removes a branch and thus decreases the effective network depth.
In the context of dataset distillation, the test network is deeper than the training network, so we can decrease the effective depth of the test network by DropPath. This approach implicitly bridges the architecture disparity between the training and test networks.
Consequently, we anticipate that DropPath will mitigate the problem of architecture overfitting on distilled datasets.

\textbf{Three-Phase Keep Rate:}
The keep rate $p$ is the key parameter that controls the effective depth of model architecture when using DropPath.
Since the mask $m = \mathtt{Bernoulli}(p)$, the effective depth gets smaller as $p$ decreases.
In the early phase of training, the model is underfitting, stochastic architecture brings optimization challenges for training the model, so we turn off DropPath by setting the keep rate $p = 1$ in the first few epochs to ensure that the network learns meaningful representations.
We then gradually decrease $p$ to decrease the effective depth and thus to decrease the architecture disparity between the effective test network and the training network until the value of $p$ reaches the predefined minimum value after several epochs.
In the final phase of training, we decrease the architecture stochasticity by increasing the value of $p$ to a higher value to ensure good training convergence.
In the experiments, we shrink the keep rate every few epochs.

\begin{algorithm}[htb]
\caption{DropPath with Three-Phase Keep Rate}\label{alg:droppath}
\begin{algorithmic}[1]
    \STATE {\bfseries Input:} the data: $\mathbf{x}$; current epoch index: $i$; decaying factor: $0<\gamma<1$; minimum keep rate: $p_{\mathrm{min}}$; final keep rate: $p_{\mathrm{final}}$; period of decay: $T$; warmup period: $W$; stabilization epoch: $S$.
    \IF{$i < W$}
        \STATE $p \gets 1$
    \ELSIF{$i<S$}
        \STATE $p \gets \mathtt{max}(p_{\mathrm{min}}, 1-\gamma\cdot \mathtt{ceil}((i - W)/T))$ 
        \COMMENT {ceil function returns the smallest integer bigger than the input}
    \ELSE
        \STATE $p \gets p_{\mathrm{final}}$
    \ENDIF
    \IF{is training}
        \STATE $m \gets \mathtt{Bernoulli}(p)$ \COMMENT{Bernoulli distribution}
        \STATE $\mathbf{y} \gets \frac{m}{p}\cdot \mathbf{x}$
    \ELSE
        \STATE $\mathbf{y} \gets \mathbf{x}$
    \ENDIF
    \STATE {\bfseries Output:} $\mathbf{y}$
\end{algorithmic}
\end{algorithm}

The pseudo-code is demonstrated in Algorithm \ref{alg:droppath}. Unless specified, we set $\gamma = 0.1$, $p_{\mathrm{min}} = 0.5$, $p_{\mathrm{final}} = 0.8$, $T = 500$, $W = 500$, $S = 3000$ in the experiments. The corresponding curve of the dynamic keep rate is shown in Figure \ref{fig:kr} of Appendix \ref{sec:app_droppath}. 

\textbf{Generalize to Single-Branch Networks:}
DropPath prunes the entire branch, so it is not applicable to single-branch networks, such as VGG~\citep{simonyan2014very}.
This is because we need to ensure the input and the output of the network are always connected, otherwise, we will obtain a trivial constant model.
By contrast, in the case of multi-branch networks such as ResNet, we prune the main path of a residual block stochastically, while the shortcut connections are always kept. 

To improve the performance of single-branch networks, we propose a variant of DropPath.
As illustrated in Figure \ref{fig:dp}(b), we add a virtual shortcut connection between two layers, such as two consecutive convolutional layers in VGG, to form a ``pseudo-residual'' block.
This structure is similar to a real residual block, however, since we are training a single-branch architecture instead of a real ResNet, the virtual shortcut connection is only used when the main path is pruned by DropPath during training.
That is to say when the main path is not pruned, the virtual shortcut connection is removed so that we are still training a single-branch network.
Correspondingly, the virtual shortcut connection is discarded during inference. It should be noted that the feature is not scaled in virtual shortcut connection. The detailed derivation is also deferred to Appendix \ref{sec:ap_scale_factor}.


\textbf{Improved Shortcut Connection:}
In the original ResNet \citep{he2016deep}, if one residual block's input shape is the same as its output shape, the shortcut connection is just an identity function, otherwise a $1\times 1$ convolution layer of a stride larger than one, which may be followed by a normalization layer as shown in Figure \ref{fig:dp}(c), is adopted in the shortcut connection to transform the input's shape to match the output's.
In the latter case, the resolution of the feature maps is divided by the stride.
For example, if the stride is $2$,  the top left entry in each $2\times2$ area of the input feature map is sampled, whereas the rest $3$ entities of the same area are directly dropped.

This naive subsampling strategy will cause dramatic information loss when we use DropPath. 
Specifically, if DropPath prunes the main path as in Figure \ref{fig:dp} (a), the shortcut connection will dominate the output of the residual block.
In this regard, the naive subsampling strategy may corrupt or degrade the quality of the features, since it always picks a fixed entry of a grid.
To tackle this issue, we replace the original shortcut connect with a $2 \times 2$ max pooling followed by a $1 \times 1$ convolutional layer with the stride of $1$.
This improved structure will preserve the most important information after pooling instead of the one from a fixed entry.
Figure \ref{fig:dp} (c) and (d) show the comparison between the original and improved shortcut connections when the shapes of input and output are different.

%

\subsection{Knowledge Distillation from Small Teacher Model}
Given sufficient training data, large models usually perform better than small models due to their larger representation capability.
Knowledge distillation aims to compress a well-trained large model (i.e., teacher model) into a smaller model (i.e., student model) without compromising too much performance.
The basic idea behind knowledge distillation is to distill the knowledge from a teacher model into a student model by forcing the student’s predictions (or internal activations) to match those of the teacher \citep{beyer2022knowledge}.
Specifically, we can use Kullback-Leibler (KL) divergence $\mathcal{L}_{KL}$ with temperature \citep{hinton2015distilling} to match the predictions of student and teacher models.
Then, we can combine the KL divergence as the regularization term in addition to the classification loss. Mathematically, the overall loss with knowledge distillation is:
\begin{equation} \label{eq:kd}
    \mathcal{L}(\mathbf{y}_s, \mathbf{y}_t, y) =  \alpha\cdot\tau^2\cdot\mathcal{L}_{KL}(\mathbf{y}_s, \mathbf{y}_t) + (1-\alpha)\cdot\mathcal{L}_{CE}(\mathbf{y}_s, y)
\end{equation}
where $\tau$ denotes the temperature factor, and $\alpha \in (0, 1)$ denotes the weight factor to balance the KL divergence $\mathcal{L}_{KL}$ and the cross-entropy loss $\mathcal{L}_{CE}$. The output logits of the student model and teacher model are denoted by $\mathbf{y}_s$ and $\mathbf{y}_t$, respectively. $y$ denotes the target.

When training on distalled dataset, small models perform better than large ones, since small models are employed as the training network to construct distilled dataset.
As a result, we adopt the small training network as the teacher model $\mathbf{y}_t$ and the large test network as the student model $\mathbf{y}_s$.
The computational overhead in knowledge distillation mainly arises from calculating $\mathbf{y}_t$, i.e., the output of the teacher model.
In this case, the computational overhead is negligible because evaluating on the small teacher model is much more efficient than on the large student model.




\subsection{Training and Data Augmentation} \label{sec:other_methods}
Besides aforementioned methods, we use the following methods to further improve the performance.

\textbf{Periodical Learning Rate: }
Because of the three-phase stepwise scheduler for the keep rate $p$, we expect the network to jump out of the current local minima, and tries to search for a better one when $p$ changes.
Inspired by \citep{huang2017snapshot}, we use a cosine annealing curve with warmup to adjust the learning rate, and we periodically reset it when $p$ changes.
Formally, the learning rate $\mathrm{lr}_i$ in the $i$-th epoch is calculated as follows:
\begin{equation} \label{eq: lr}
    \mathrm{lr}_i = 
    \begin{cases}
         \lambda_i \cdot \frac{\mathrm{mod}(i, t)}{T_{\mathrm{warm}}} \cdot \mathrm{lr}_{\mathrm{max}} , ~~~~~~~~~~~~\text{if $\mathrm{mod}(i, t)\leq T_{\mathrm{warm}}$}, \\
        0.5\lambda_i(1+\mathrm{cos}(\pi\frac{\mathrm{mod}(i, t) - T_{\mathrm{warm}}}{T_{\mathrm{max}}-T_{\mathrm{warm}}}))\cdot \mathrm{lr}_{\mathrm{max}},~ \text{otherwise}.
    \end{cases}
\end{equation}
where $T$ is the decay period of the keep rate $p$ of DropPath, $S$ is the stabilization epoch. $t=T$ when $i<S$, otherwise $t=S$. $\lambda_i = \lambda^{\lfloor \mathtt{min}(i, S)/T \rfloor}$ where $\lambda$ is a base decaying factor, and $\lfloor\cdot\rfloor$ denotes the floor function. $\mathrm{lr}_{\mathrm{max}}$ denotes the maximum learning rate, $\mathrm{mod}(x,y)$ denotes the remainder of $x/y$. The maximum iterations of the cosine annealing function and the number of warmup epochs are denoted by $T_{\mathrm{max}}$ and $T_{\mathrm{warm}}$, respectively. Figure \ref{fig:period_lr} of Appendix \ref{sec:app_droppath} shows an example of how the learning rate changes.

\textbf{Better Optimizer: }
Lion \citep{chen2023symbolic} is a gradient symbol-based optimizer. It has faster convergence speed and is capable of finding better local minima for ResNets. Thus, we use Lion as the default optimizer in our experiments.

\textbf{Stronger Augmentation: }
The data augmentation strategy used in MTT~\citep{cazenavette2022dataset} samples a single augmentation operation from a pool to augment the input image.
However, we observe that sampling more operations will better diversify the model's inputs and thus improve the performance, especially when IPC is small.
For convenience, when sampling $k$ operations, we call this strategy $k$-fold augmentation.
Empirically, we use 2-fold augmentation when IPC is $10$ or $50$ and 4-fold augmentation when IPC is $1$.

In summary, our proposed methods share a common characteristic of smoothing the optimization problem that can improve generalization: \textbf{(a)} in terms of architecture, DropPath smooths the predictions by forming an implicit ensemble of sub-networks; \textbf{(b)} knowledge distillation smooths the objective function by introducing the predictions of teacher models as soft labels; \textbf{(c)} better optimizer is capable of finding flatter local minima; \textbf{(d)} stronger data augmentation smooth the loss landscape in the sample space \citep{shorten2019survey, rebuffi2021data}.
\section{Experiments}
In this section, we evaluate our method on different dataset distillation algorithms, different numbers of instances per class (IPC), different datasets and different network architectures.
Our methods are shown effective in mitigating architecture overfitting in these settings and generic to improve the performance on limited real data. In addition, we plot the Hessian eigenvalues and visualize the landscape of different models to corroborate the smoothing effect of the proposed methods. Ultimately, we conduct extensive ablation studies for analysis. Implementation details are deferred to Appendix \ref{sec:app_imple}.

\subsection{Mitigate Architecture Overfitting in Dateset Distillation} \label{sec:4.1}

\begin{table}[htb]
    \centering
    \caption{Experimental settings. \textit{DP} denotes DropPath with three-phase keep rate, \textit{KD} denotes knowledge distillation. Besides, the miscellaneous (Misc.) includes the methods in Section~\ref{sec:other_methods}.}
    \vspace{0.2cm}
    \begin{tabular}{c|c c c}
    \toprule[1.5pt]
        Method & DP & KD & Misc. \\
        \midrule[1pt]
        Baseline & \redcross & \redcross & \redcross\\
        w/o DP\&KD & \redcross & \redcross & \greencheck\\
         w/o DP & \redcross & \greencheck & \greencheck\\
         w/o KD & \greencheck & \redcross & \greencheck\\
        Full & \greencheck & \greencheck & \greencheck\\
    \bottomrule[1.5pt]    
    \end{tabular}
    \label{tab:exp_setting}
\end{table}

\begin{table*}[ht]
    \centering
    \caption{Test accuracies of models trained on the distilled data of \textbf{CIFAR10} and \textbf{CIFAR100} \citep{krizhevsky2009learning} with different IPCs. 3-layer CNN is the architecture used for data distillation and is the teacher model of knowledge distillation. The results in the bracket indicate the gaps from the baseline performance of 3-layer CNN. Note that for IPC=100/500, the teacher model of ResNet50 is ResNet18 w/o DP\&KD. The results in bold are the best results among different settings. Note that DP and KD are not applicable for 3-layer CNN, so we do not have the test accuracy of 3-layer CNN in these settings.}
    \vspace{-1em}
    \tabcolsep=0.36em
    \scriptsize
    \subtable[CIFAR10]{
    \begin{tabular}{c|c|c|c|c c c c}
        \toprule[1.5pt]
         DD & IPC & Methods & \makecell{3-layer\\CNN} & ResNet18 & AlexNet & VGG11 & ResNet50\\
         \midrule[1pt]
          \multirow{15}{*}{\rotatebox{90}{FRePo \citep{zhou2022dataset}}}& \multirow{5}{*}{1} & Baseline & 44.3 & 34.4 \fs{(\red{-9.9})} & 41.8 \fs{(\red{-2.5})} & 44.0 \fs{(\red{-0.3})} & 25.9 \fs{(\red{-18.4})}\\
         ~ & ~ & w/o DP\&KD & \textbf{44.8} \fs{(\blue{+0.5})} & 35.6 \fs{(\red{-8.7})} & 47.4 \fs{(\blue{+3.1})} & 41.5 \fs{(\red{-2.8})} & 30.3 \fs{(\red{-14.0})}\\
          ~ & ~ & w/o DP & - & 47.2 \fs{(\blue{+2.9})} & 49.7 \fs{(\blue{+5.4})} & 48.7 \fs{(\blue{+4.4})} & 39.3 \fs{(\red{-5.0})}\\
         ~ & ~ & w/o KD & - & 37.0 \fs{(\red{-7.3})} & 46.0 \fs{(\blue{+1.7})} & 41.1 \fs{(\red{-3.2})} & 32.5 \fs{(\red{-11.8})}\\
        ~ & ~ & \cellcolor{gray!40} \textbf{Full} & \cellcolor{gray!40}- & \cellcolor{gray!40} \textbf{49.3} \fs{(\blue{+5.0})} & \cellcolor{gray!40}\textbf{50.7} \fs{(\blue{+6.4})} & \cellcolor{gray!40}\textbf{48.8} \fs{(\blue{+4.5})} & \cellcolor{gray!40} \textbf{41.5} \fs{(\red{-2.8})}\\  
         \cmidrule{2-8}
         ~ & \multirow{5}{*}{10} & Baseline & 63.0 & 55.6 \fs{(\red{-7.4})} & 59.3 \fs{(\red{-3.6})}& 61.3 \fs{(\red{-1.7})}& 44.4 \fs{(\red{-18.6})}\\
          ~ & ~ & w/o DP\&KD & \textbf{64.7} \fs{(\blue{+1.7})}& 61.0 \fs{(\red{-2.0})} & 62.3 \fs{(\red{-0.7})} & 62.4 \fs{(\red{-0.6})} & 54.7 \fs{(\red{-8.3})}\\
          ~ & ~ & w/o DP & - & 64.0 \fs{(\blue{+1.0})} & 63.3 \fs{(\blue{+0.3})} & 63.6 \fs{(\blue{+0.6})} & 57.7 \fs{(\red{-5.3})}\\
          ~ & ~ & w/o KD & - & 63.9 \fs{(\blue{+0.9})} & 63.8 \fs{(\blue{+0.8})}& 62.2 \fs{(\red{-0.8})}& 54.0 \fs{(\red{-9.0})}\\
         ~ & ~ & \cellcolor{gray!40}\textbf{Full} & \cellcolor{gray!40}- & \cellcolor{gray!40}\textbf{66.6} \fs{(\blue{+3.6})} & \cellcolor{gray!40}\textbf{64.8} \fs{(\blue{+1.8})} & \cellcolor{gray!40}\textbf{65.4} \fs{(\blue{+2.4})}& \cellcolor{gray!40}\textbf{62.4} \fs{(\red{-0.6})}\\
         \cmidrule{2-8}
         ~& \multirow{5}{*}{50} & Baseline & 70.5 & 66.7 \fs{(\red{-3.8})} & 66.8 \fs{(\red{-3.7})} & 68.3 \fs{(\red{-2.2})} & 60.5 \fs{(\red{-10.0})}\\
          ~ & ~ & w/o DP\&KD & \textbf{72.4} \fs{(\blue{+1.9})} & 73.0 \fs{(\blue{+2.5})} & 71.0 \fs{(\blue{+0.5})} & 70.9 \fs{(\blue{+0.4})}& 71.2 \fs{(\blue{+0.7})}\\
         ~ & ~ & w/o DP & - & 73.9 \fs{(\blue{+3.4})} & 72.1 \fs{(\blue{+1.6})} & 72.0 \fs{(\blue{+1.5})} & 72.9 \fs{(\blue{+2.4})}\\
         ~ & ~ & w/o KD & - & 74.5 \fs{(\blue{+4.0})} & 71.5 \fs{(\blue{+1.0})} & 70.1 \fs{(\red{-0.4})} & 70.6 \fs{(\blue{+0.1})}\\
         ~ & ~ &  \cellcolor{gray!40}\textbf{Full} & \cellcolor{gray!40}- & \cellcolor{gray!40}\textbf{74.5} \fs{(\blue{+4.0})} & \cellcolor{gray!40}\textbf{73.2} \fs{(\blue{+2.7})}& \cellcolor{gray!40}\textbf{72.8} \fs{(\blue{+2.3})}& \cellcolor{gray!40}\textbf{73.2} \fs{(\blue{+2.7})}\\
         \midrule[1pt]
         \multirow{15}{*}{\rotatebox{90}{MTT \citep{cazenavette2022dataset}}}& \multirow{5}{*}{1} & Baseline & \textbf{48.3} & 37.2 \fs{(\red{-11.1})} & 40.5 \fs{(\red{-7.8})} & 39.3 \fs{(\red{-9.0})}& 22.4 \fs{(\red{-25.9})}\\
         ~ & ~ & w/o DP\&KD & 46.8 \fs{(\red{-1.5})} & 36.9 \fs{(\red{-11.4})} & 43.2 \fs{(\red{-5.1})} & 36.7 \fs{(\red{-11.6})} & 24.7 \fs{(\red{-23.6})}\\
         ~ & ~ & w/o DP & - & 41.6 \fs{(\red{-6.7})} & 46.7 \fs{(\red{-1.6})}& 38.6 \fs{(\red{-9.7})}& 32.4 \fs{(\red{-15.9})}\\
         ~ & ~ & w/o KD & - &  35.5 \fs{(\red{-12.8})} & 41.1 \fs{(\red{-7.2})} & 34.4 \fs{(\red{-13.9})} & 28.5 \fs{(\red{-19.8})}\\
         ~ & ~ &\cellcolor{gray!40} \textbf{Full} & \cellcolor{gray!40}- & \cellcolor{gray!40}\textbf{47.2} \fs{(\red{-1.1})} & \cellcolor{gray!40}\textbf{47.3} \fs{(\red{-1.0})}& \cellcolor{gray!40}\textbf{44.1} \fs{(\red{-4.2})} & \cellcolor{gray!40}\textbf{43.0} \fs{(\red{-5.3})}\\
         \cmidrule{2-8}
         ~& \multirow{5}{*}{10} & Baseline & 63.6 & 48.9 \fs{(\red{-14.7})} & 56.9 \fs{(\red{-6.7})} & 52.6 \fs{(\red{-11.0})}& 28.1 \fs{(\red{-35.5})}\\
         ~ & ~ & w/o DP\&KD & \textbf{65.0} \fs{(\blue{+1.4})} & 51.3 \fs{(\red{-12.3})} & 60.7 \fs{(\red{-2.9})} & 56.0 \fs{(\red{-7.6})} & 39.8 \fs{(\red{-23.8})}\\
         ~ & ~ & w/o DP & - & 61.4 \fs{(\red{-2.2})} & 52.7 \fs{(\red{-10.9})} & 48.8 \fs{(\red{-14.8})} & 49.9 \fs{(\red{-13.7})}\\
         ~ & ~ & w/o KD & - & 60.7 \fs{(\red{-2.9})} & 59.2 \fs{(\red{-4.4})} & 57.6 \fs{(\red{-6.0})} & 47.5 \fs{(\red{-16.1})}\\
         ~ & ~ & \cellcolor{gray!40}\textbf{Full} & \cellcolor{gray!40}- & \cellcolor{gray!40}\textbf{67.4} \fs{(\blue{+3.8})} & \cellcolor{gray!40}\textbf{68.3} \fs{(\blue{+4.7})}& \cellcolor{gray!40}\textbf{67.1} \fs{(\blue{+3.5})}& \cellcolor{gray!40}\textbf{63.8} \fs{(\blue{+0.2})}\\
         \cmidrule{2-8}
         ~& \multirow{5}{*}{50} & Baseline & 70.2 & 62.3 \fs{(\red{-7.9})} & 67.5 \fs{(\red{-2.7})} & 63.0 \fs{(\red{-7.2})} & 53.1 \fs{(\red{-17.1})}\\
          ~ & ~ & w/o DP\&KD & \textbf{70.5} \fs{(\blue{+0.3})} & 68.1 \fs{(\red{-2.1})} & 69.5 \fs{(\red{-0.7})} & 67.6 \fs{(\red{-2.6})}& 66.5 \fs{(\red{-3.7})}\\
          ~ & ~ & w/o DP & - & 66.9 \fs{(\red{-3.3})}& 63.8 \fs{(\red{-6.4})}& 61.2 \fs{(\red{-9.0})}& 66.8 \fs{(\red{-3.4})}\\
          ~ & ~ & w/o KD & - & 69.8 \fs{(\red{-0.4})} & 67.2 \fs{(\red{-3.0})} & 69.0 \fs{(\red{-1.2})} & 65.0 \fs{(\red{-5.2})}\\
         ~ & ~ & \cellcolor{gray!40}\textbf{Full} & \cellcolor{gray!40}- & \cellcolor{gray!40}\textbf{71.0} \fs{(\blue{+0.8})} & \cellcolor{gray!40}\textbf{72.0} \fs{(\blue{+1.8})} & \cellcolor{gray!40}\textbf{69.5} \fs{(\red{-1.2})}& \cellcolor{gray!40}\textbf{70.0} \fs{(\red{-0.2})}\\
         \midrule[1pt]
        \multirow{15}{*}{\rotatebox{90}{DATM \cite{guotowards}}}& \multirow{5}{*}{10} & Baseline & 58.2 & 50.4 \fs{(\red{-7.8})} & 58.4 \fs{(\blue{+0.2})} & 53.1 \fs{(\red{-5.1})} & 28.8 \fs{(\red{-29.4})}\\
        ~ & ~ & w/o DP\&KD & 66.5 \fs{(\blue{+8.3})} & 51.0 \fs{(\red{-7.2})} & 60.3 \fs{(\blue{+5.1})} & 57.4 \fs{(\red{-0.8})} & 39.6 \fs{(\red{-18.6})}\\
        ~ & ~ & w/o DP & - & 54.9 \fs{(\red{-3.3})} & 65.8 \fs{(\blue{+7.6})} & 61.9 \fs{(\blue{+3.7})} & 43.2 \fs{(\red{-15.0})}\\
        ~ & ~ & w/o KD & - & 59.6 \fs{(\blue{+1.4})} & 63.5 \fs{(\blue{+5.3})} & 60.5 \fs{(\blue{+2.3})} & 53.5 \fs{(\red{-4.7})}\\
        ~ & ~ & \cellcolor{gray!40} \textbf{Full} & \cellcolor{gray!40}- & \cellcolor{gray!40} \textbf{64.3} \fs{(\blue{+6.1})} & \cellcolor{gray!40}\textbf{67.5} \fs{(\blue{+9.3})} & \cellcolor{gray!40}\textbf{63.4} \fs{(\blue{+5.2})} & \cellcolor{gray!40} \textbf{59.8} \fs{(\blue{+1.6})}\\ 
        \cmidrule{2-8}
        ~ & \multirow{5}{*}{50} & Baseline & 70.0 & 69.2 \fs{(\red{-0.8})} & 71.5 \fs{(\blue{+1.5})}& 66.9 \fs{(\red{-3.1})}& 54.4 \fs{(\red{-15.6})}\\
        ~ & ~ & w/o DP\&KD & 74.5  \fs{(\blue{+4.5})} & 72.1 \fs{(\blue{+2.1})} & 73.7 \fs{(\blue{+3.7})} & 71.8 \fs{(\blue{+1.8})} & 70.0 \fs{(\blue{+0.0})}\\
        ~ & ~ & w/o DP & - & 73.0 \fs{(\blue{+3.0})} & 75.3 \fs{(\blue{+5.3})} & 71.9 \fs{(\blue{+1.9})} & 72.5 \fs{(\blue{+2.5})}\\
        ~ & ~ & w/o KD & - & 75.4 \fs{(\blue{+5.4})} & 73.8 \fs{(\blue{+3.8})} & 73.1 \fs{(\blue{+3.1})} & 73.3 \fs{(\blue{+3.3})}\\
        ~ & ~ & \cellcolor{gray!40}\textbf{Full} & \cellcolor{gray!40}- & \cellcolor{gray!40}\textbf{75.7} \fs{(\blue{+5.7})} & \cellcolor{gray!40}\textbf{77.2} \fs{(\blue{+7.2})} & \cellcolor{gray!40}\textbf{74.7} \fs{(\blue{+4.7})}& \cellcolor{gray!40}\textbf{75.7} \fs{(\blue{+5.7})}\\ 
        \cmidrule{2-8}
        ~ & \multirow{5}{*}{500} & Baseline & 76.5 & 82.5 \fs{(\blue{+6.0})} & 80.1 \fs{(\blue{+3.6})}& 77.0 \fs{(\blue{+0.5})}& 79.8 \fs{(\blue{+3.3})}\\
        ~ & ~ & w/o DP\&KD & 83.3  \fs{(\blue{+6.8})} & 86.0 \fs{(\blue{+9.5})} & 83.5 \fs{(\blue{+7.0})} & 82.3 \fs{(\blue{+5.8})} & 85.8 \fs{(\blue{+9.3})}\\
        ~ & ~ & w/o DP & - & 85.9 \fs{(\blue{+9.4})} & 84.4 \fs{(\blue{+7.9})} & 83.6 \fs{(\blue{+7.1})} & 86.5 \fs{(\blue{+10.0})}\\
        ~ & ~ & w/o KD & - & 86.7 \fs{(\blue{+10.2})} & 84.3 \fs{(\blue{+7.8})} & 84.2 \fs{(\blue{+7.7})} & 87.2 \fs{(\blue{+10.7})}\\
        ~ & ~ & \cellcolor{gray!40}\textbf{Full} & \cellcolor{gray!40}- & \cellcolor{gray!40}\textbf{86.8} \fs{(\blue{+10.3})} & \cellcolor{gray!40}\textbf{85.1} \fs{(\blue{+8.6})} & \cellcolor{gray!40}\textbf{85.3} \fs{(\blue{+8.8})}& \cellcolor{gray!40}\textbf{87.9} \fs{(\blue{+11.4})}\\ 
         \bottomrule[1.5pt]
    \end{tabular}
    \label{tab:cifar10}
    }
    \subtable[CIFAR100]{
       \begin{tabular}{c|c|c|c|c c c c}
        \toprule[1.5pt]
         DD & IPC & Methods & \makecell{3-layer\\CNN} & ResNet18 & AlexNet & VGG11 & ResNet50\\
         \midrule[1pt]
          \multirow{15}{*}{\rotatebox{90}{FRePo \citep{zhou2022dataset}}}& \multirow{5}{*}{1} & Baseline & \textbf{26.2} & 18.7 \fs{(\red{-7.5})} & 22.9 \fs{(\red{-3.3})} &  22.6 \fs{(\red{-3.6})} & 13.5 \fs{(\red{-12.7})}\\
         ~ & ~ & w/o DP\&KD & 26.1  \fs{(\red{-0.1})} & 16.0 \fs{(\red{-10.2})} & 22.3 \fs{(\red{-3.9})} & 18.4 \fs{(\red{-7.8})} & 14.5 \fs{(\red{-11.7})}\\
          ~ & ~ & w/o DP & - & 21.3 \fs{(\red{-4.9})} & 23.9 \fs{(\red{-2.3})} & 21.8 \fs{(\red{-4.4})} & 18.2 \fs{(\red{-8.0})}\\
         ~ & ~ & w/o KD & - & 17.1 \fs{(\red{-9.1})} & 22.1 \fs{(\red{-4.1})} & 17.9 \fs{(\red{-8.3})} & 14.3 \fs{(\red{-11.9})}\\
         ~ & ~ & \cellcolor{gray!40} \textbf{Full} & \cellcolor{gray!40}- & \cellcolor{gray!40} \textbf{24.4} \fs{(\red{-1.8})} & \cellcolor{gray!40}\textbf{25.3} \fs{(\red{-0.9})} & \cellcolor{gray!40}\textbf{24.0} \fs{(\red{-2.2})} & \cellcolor{gray!40} \textbf{23.7} \fs{(\red{-2.5})}\\ 
         \cmidrule{2-8}
         ~ & \multirow{5}{*}{10} & Baseline & 34.4 & 32.1 \fs{(\red{-2.3})} &  33.1 \fs{(\red{-1.3})}&  34.1 \fs{(\red{-0.3})}&  28.1\fs{(\red{-6.3})}\\
          ~ & ~ & w/o DP\&KD & \textbf{40.2} \fs{(\blue{+5.8})} & 35.3 \fs{(\blue{+0.9})} & 37.9 \fs{(\blue{+3.5})} & 37.2 \fs{(\blue{+2.8})} & 33.7 \fs{(\red{-0.7})}\\
          ~ & ~ & w/o DP & - & 39.4 \fs{(\blue{+5.0})} & 39.2 \fs{(\blue{-4.8})} & 38.9 \fs{(\blue{+4.5})} & 38.5 \fs{(\blue{+4.1})}\\
          ~ & ~ & w/o KD & - & 34.8 \fs{(\blue{+0.4})} & 38.5 \fs{(\blue{+4.1})}& 36.6 \fs{(\blue{+2.2})}&  35.0\fs{(\blue{+0.6})}\\
         ~ & ~ & \cellcolor{gray!40}\textbf{Full} & \cellcolor{gray!40}- & \cellcolor{gray!40}\textbf{40.6} \fs{(\blue{+6.2})} & \cellcolor{gray!40}\textbf{39.9} \fs{(\blue{+5.5})} & \cellcolor{gray!40}\textbf{39.4} \fs{(\blue{+5.0})}& \cellcolor{gray!40}\textbf{40.1} \fs{(\blue{+5.7})}\\
         \cmidrule{2-8}
         ~& \multirow{5}{*}{50} & Baseline & 42.1 & 46.7 \fs{(\blue{+4.6})} & 45.5 \fs{(\blue{+3.4})} & 45.5 \fs{(\blue{+3.4})} & 45.8 \fs{(\blue{+3.7})}\\
          ~ & ~ & w/o DP\&KD & \textbf{46.2} \fs{(\blue{+4.1})}& 46.8 \fs{(\blue{+4.7})} & 46.1 \fs{(\blue{+4.0})} & 45.5 \fs{(\blue{+3.4})}& 46.9 \fs{(\blue{+4.8})}\\
         ~ & ~ & w/o DP & - & 48.3 \fs{(\blue{+6.2})} & 44.6 \fs{(\blue{+2.5})} & 45.8 \fs{(\blue{+3.7})} & 48.7 \fs{(\blue{+6.6})}\\
         ~ & ~ & w/o KD & - & 47.2 \fs{(\blue{+5.1})} & \textbf{47.0} \fs{(\blue{+4.9})} & 45.0 \fs{(\blue{+2.9})} & 46.1 \fs{(\blue{+4.0})}\\
         ~ & ~ &  \cellcolor{gray!40}\textbf{Full} & \cellcolor{gray!40}- & \cellcolor{gray!40}\textbf{48.5} \fs{(\blue{+6.4})} & \cellcolor{gray!40} 46.6 \fs{(\blue{+4.5})}& \cellcolor{gray!40}\textbf{46.7} \fs{(\blue{+4.6})}& \cellcolor{gray!40}\textbf{49.1} \fs{(\blue{+7.0})}\\
         \midrule[1pt]
         \multirow{15}{*}{\rotatebox{90}{MTT \citep{cazenavette2022dataset}}}& \multirow{5}{*}{1} & Baseline & 24.4 & 14.3 \fs{(\red{-10.1})} & 17.0 \fs{(\red{-7.4})} & 15.6 \fs{(\red{-8.8})}& 4.6 \fs{(\red{-19.8})}\\
         ~ & ~ & w/o DP\&KD & \textbf{25.0} \fs{(\blue{+0.6})} & 12.5 \fs{(\red{-11.9})} & 20.6 \fs{(\red{-3.8})} & 8.2 \fs{(\red{-16.2})} & 6.0 \fs{(\red{-18.4})}\\
         ~ & ~ & w/o DP & - & 13.3 \fs{(\red{-11.1})} & 24.4 \fs{(\blue{+0.0})}& 10.2 \fs{(\red{-14.2})}&  8.5\fs{(\red{-15.9})}\\
         ~ & ~ & w/o KD & - &  13.6 \fs{(\red{-10.8})} & 19.7 \fs{(\red{-4.7})} & 12.4 \fs{(\red{-12.0})} & 9.3 \fs{(\red{-15.1})}\\
         ~ & ~ &\cellcolor{gray!40} \textbf{Full} & \cellcolor{gray!40}- & \cellcolor{gray!40}\textbf{24.9} \fs{(\blue{+0.5})} & \cellcolor{gray!40}\textbf{25.8} \fs{(\blue{+1.4})}& \cellcolor{gray!40}\textbf{22.1} \fs{(\red{-2.3})} & \cellcolor{gray!40}\textbf{24.6} \fs{(\blue{+0.2})}\\
         \cmidrule{2-8}
         ~& \multirow{5}{*}{10} & Baseline & 38.4 & 32.9 \fs{(\red{-5.5})} & 33.7 \fs{(\red{-4.7})} & 28.8 \fs{(\red{-9.6})}& 22.5 \fs{(\red{-15.9})}\\
         ~ & ~ & w/o DP\&KD & \textbf{38.5}\fs{(\blue{+0.1})} & 32.7 \fs{(\red{-5.7})} & 36.0 \fs{(\red{-2.4})} & 33.9 \fs{(\red{-4.5})} & 30.6 \fs{(\red{-7.8})}\\
         ~ & ~ & w/o DP & - & 35.0 \fs{(\red{-3.4})} & 38.2 \fs{(\red{-0.2})} & 35.5 \fs{(\red{-2.9})} & 34.2 \fs{(\red{-4.2})}\\
         ~ & ~ & w/o KD & - & 34.6 \fs{(\red{-3.8})} & 34.9 \fs{(\red{-3.5})} & 33.2 \fs{(\red{-5.2})} & 32.9 \fs{(\red{-5.5})}\\
         ~ & ~ & \cellcolor{gray!40}\textbf{Full} & \cellcolor{gray!40}- & \cellcolor{gray!40}\textbf{38.4} \fs{(\blue{+0.0})} & \cellcolor{gray!40}\textbf{39.9} \fs{(\blue{+1.5})}& \cellcolor{gray!40}\textbf{36.4} \fs{(\red{-2.0})}& \cellcolor{gray!40}\textbf{38.5} \fs{(\blue{+0.1})}\\
         \cmidrule{2-8}
         ~& \multirow{5}{*}{50} & Baseline & 44.5 &  43.1 \fs{(\red{-1.4})} & 41.4 \fs{(\red{-3.1})} & 39.3 \fs{(\red{-5.2})} & 38.7 \fs{(\red{-5.8})}\\
          ~ & ~ & w/o DP\&KD & \textbf{46.0} \fs{(\blue{+1.5})} & 46.2 \fs{(\blue{+1.7})} & 46.1 \fs{(\blue{+1.6})} & 44.5 \fs{(\blue{+0.0})}& 45.5 \fs{(\blue{+1.0})}\\
          ~ & ~ & w/o DP & - & 47.2 \fs{(\blue{+2.7})}& 47.1 \fs{(\blue{+2.6})}& 45.1 \fs{(\blue{+0.6})}& 47.2 \fs{(\blue{+2.7})}\\
          ~ & ~ & w/o KD & - & 46.9 \fs{(\blue{+2.4})} & 45.7 \fs{(\blue{+1.2})} & 43.4 \fs{(\red{-1.1})} & 46.8 \fs{(\blue{+2.3})}\\
         ~ & ~ & \cellcolor{gray!40}\textbf{Full} & \cellcolor{gray!40}- & \cellcolor{gray!40}\textbf{48.9} \fs{(\blue{+4.4})} & \cellcolor{gray!40}\textbf{47.6} \fs{(\blue{+3.1})} & \cellcolor{gray!40}\textbf{45.1} \fs{(\blue{+0.6})}& \cellcolor{gray!40}\textbf{49.4} \fs{(\blue{+4.9})}\\
                 \midrule[1pt]
        \multirow{15}{*}{\rotatebox{90}{DATM \cite{guotowards}}}& \multirow{5}{*}{10} & Baseline & 29.6 & 21.9 \fs{(\red{-7.7})} & 26.8 \fs{(\red{-2.8})} & 21.2 \fs{(\red{-8.4})} & 8.7 \fs{(\red{-20.9})}\\
        ~ & ~ & w/o DP\&KD & 32.4 \fs{(\blue{+2.8})} & 25.5 \fs{(\red{-4.1})} & 32.4 \fs{(\blue{+2.8})} & 26.4 \fs{(\red{-3.2})} & 17.9 \fs{(\red{-11.7})}\\
        ~ & ~ & w/o DP & - & 29.7 \fs{(\blue{+0.1})} & 34.6 \fs{(\blue{+5.0})} & 31.1 \fs{(\blue{+1.5})} & 24.2 \fs{(\red{-5.4})}\\
        ~ & ~ & w/o KD & - & 29.5 \fs{(\red{-0.1})} & 32.3 \fs{(\blue{+2.7})} & 30.6 \fs{(\blue{+1.0})} & 28.2 \fs{(\red{-1.4})}\\
        ~ & ~ & \cellcolor{gray!40} \textbf{Full} & \cellcolor{gray!40}- & \cellcolor{gray!40} \textbf{33.9} \fs{(\blue{+4.3})} & \cellcolor{gray!40}\textbf{35.3} \fs{(\blue{+5.7})} & \cellcolor{gray!40}\textbf{33.1} \fs{(\blue{+3.5})} & \cellcolor{gray!40} \textbf{35.2} \fs{(\blue{+5.6})}\\ 
        \cmidrule{2-8}
        ~ & \multirow{5}{*}{50} & Baseline & 46.8 & 44.0 \fs{(\red{-2.8})} & 44.7 \fs{(\red{-2.1})}& 41.7 \fs{(\red{-5.1})}& 39.1 \fs{(\red{-7.7})}\\
        ~ & ~ & w/o DP\&KD & 47.4 \fs{(\blue{+0.6})} & 47.6 \fs{(\blue{+0.8})} & 48.6 \fs{(\blue{+1.8})} & 46.5 \fs{(\red{-0.3})} & 46.6 \fs{(\red{-0.2})}\\
        ~ & ~ & w/o DP & - & 51.3 \fs{(\blue{+4.5})} & 50.6 \fs{(\blue{+3.8})} & 48.8 \fs{(\blue{+2.0})} & 50.5 \fs{(\blue{+3.7})}\\
        ~ & ~ & w/o KD & - & 49.9 \fs{(\blue{+3.4})} & 48.0 \fs{(\blue{+1.2})} & 48.2 \fs{(\blue{+1.4})} & 50.7 \fs{(\blue{+3.9})}\\
        ~ & ~ & \cellcolor{gray!40}\textbf{Full} & \cellcolor{gray!40}- & \cellcolor{gray!40}\textbf{52.0} \fs{(\blue{+5.2})} & \cellcolor{gray!40}\textbf{50.6} \fs{(\blue{+3.8})} & \cellcolor{gray!40}\textbf{50.3} \fs{(\blue{+3.5})}& \cellcolor{gray!40}\textbf{54.0} \fs{(\blue{+7.2})}\\ 
        \cmidrule{2-8}
        ~ & \multirow{5}{*}{100} & Baseline & 52.5 & 55.0 \fs{(\blue{+2.5})} & 53.1 \fs{(\blue{+0.6})}& 51.0 \fs{(\red{-1.5})}& 52.1 \fs{(\red{-0.4})}\\
        ~ & ~ & w/o DP\&KD & 53.6 \fs{(\blue{+1.1})} & 58.4 \fs{(\blue{+5.9})} & 55.6 \fs{(\blue{+3.1})} & 55.1 \fs{(\blue{+2.6})} & 58.9 \fs{(\blue{+6.4})}\\
        ~ & ~ & w/o DP & - & 59.3 \fs{(\blue{+6.8})} & 56.7 \fs{(\blue{+4.2})} & 56.3 \fs{(\blue{+3.8})} & 59.7 \fs{(\blue{+7.2})}\\
        ~ & ~ & w/o KD & - & 60.5 \fs{(\blue{+8.0})} & 55.8 \fs{(\blue{+3.3})} & 57.2 \fs{(\blue{+4.7})} & 60.6 \fs{(\blue{+8.1})}\\
        ~ & ~ & \cellcolor{gray!40}\textbf{Full} & \cellcolor{gray!40}- & \cellcolor{gray!40}\textbf{60.5} \fs{(\blue{+8.0})} & \cellcolor{gray!40}\textbf{56.5} \fs{(\blue{+4.0})} & \cellcolor{gray!40}\textbf{58.0} \fs{(\blue{+5.5})}& \cellcolor{gray!40}\textbf{60.9} \fs{(\blue{+8.4})}\\ 
         \bottomrule[1.5pt]
    \end{tabular}
    \label{tab:cifar100}
    }
    \vspace{-1.5em}
\end{table*}

We first evaluate our method on three representative dataset distillation (DD) algorithms, i.e., neural Feature Regression with Pooling (FRePo) \citep{zhou2022dataset}, Matching Training Trajectories (MTT) \citep{cazenavette2022dataset} and Difficulty-Aligned Trajectory Matching (DATM) \cite{guotowards}.
Furthermore, we test several ablations of our methods, the names and the settings of each ablation are elaborated in Table \ref{tab:exp_setting}. 

We comprehensively evaluate the performance of these methods under various settings, including different numbers of instances per class (IPC), different datasets and different architectures of the test networks.
Table \ref{tab:cifar10} demonstrate the results on CIFAR10, and the results on CIFAR100 and Tiny-ImageNet are reported in Table~\ref{tab:cifar100} and Table~\ref{tab:timage}, respectively.
Note that, DropPath and knowledge distillation are not applicable when we use the same architecture for training and test networks, i.e., 3-layer CNN, because 1) it is too shallow for DropPath; 2) we will converge to the teacher model if we use the same model architecture for the teacher and the student models.
We can observe from these results that architecture overfitting is more severe in the case of small IPCs and large architecture discrepancy between the training networks and the test networks, but both DropPath and knowledge distillation is capable of mitigating it.
In addition, combining them can further improve the performance and overcome architecture overfitting in many cases.
For instance, when evaluating our method on distilled images of MTT (CIFAR10, IPC=10), it contributes performance gains of $18.5\%$ and $35.7\%$ for ResNet18 and ResNet50, respectively. We are also interested in how much performance gap between training and test networks we can close. Surprisingly, when IPC=$10$ and $50$, the test accuracies of most network architectures surpass that of the architecture identical to the training network. Along with it, the gaps between different test networks, such as ResNet18 and ResNet50, are also narrowed down in most cases. Additionally, when the IPC reaches 500, our method can still contribute to performance gain.

DropPath enables an implicit ensemble of the shallow subnetworks and thus mitigates architecture overfitting. However, each of these sub-networks may have sub-optimal performance. Knowledge distillation can address this issue by encouraging similar outputs between the teacher model and the sub-networks and thus further improves the performance. By contrast, the contribution of knowledge distillation could be marginal without DropPath due to the big difference in architecture \citep{mirzadeh2020improved}.
Empirically, combining DropPath with knowledge distillation not only achieves the best performance, but also greatly decreases the performance difference among different test network architectures.

To better validate the effectiveness of our method, we report the standard deviations of test accuracies of CIFAR10 (FRePo) in Table \ref{tab:app_std}. We calculate these standard deviations by running the experiments three times with different random seeds. It can be observed that the standard deviation generally increases as IPC decreases. The reason could be that when IPC gets smaller, there are more solutions that make the training error zero, so the performance of training becomes more sensitive to initialization. Despite this, we can still see significant improvement introduced by our methods.

\begin{table}[ht]
\vspace{-1em}
    \centering
    \caption{The average test accuracies of models trained on the distilled data of \textbf{CIFAR10} \citep{krizhevsky2009learning} with different IPCs. The number after $\pm$ denotes the standard deviation. These results are obtained through three repetitive experiments with different random seeds. 3-layer CNN is the architecture used in distillation and is the teacher model of knowledge distillation.}
    \scriptsize
    \tabcolsep=0.5em
    \begin{tabular}{c|c|c|c c c c}
        \toprule[1.5pt]
         DD & IPC & Methods & ResNet18 & AlexNet & VGG11 & ResNet50\\
         \midrule[1pt]
          \multirow{7}{*}{\rotatebox{90}{FRePo}}& \multirow{5}{*}{1} & w/o DP\&KD & 35.6 \fs{$\pm 2.5$} & 47.4 \fs{$\pm 0.9$} & 41.5 \fs{$\pm 1.1$} & 30.3 \fs{$\pm 1.9$}\\
          ~ & ~ & w/o DP & 47.2 \fs{$\pm 0.5$} & 49.7 \fs{$\pm 0.7$} &48.7 \fs{$\pm 0.6$} & 39.3 \fs{$\pm 1.4$}\\
         ~ & ~ & w/o KD & 37.0 \fs{$\pm 1.0$} & 46.0 \fs{$\pm 0.6$} & 41.1 \fs{$\pm 1.3$} & 32.5 \fs{$\pm 1.4$}\\
        ~ & ~ &  Full & 49.3 \fs{$\pm 0.6$} & 50.7 \fs{$\pm 0.1$} & 48.8 \fs{$\pm 0.4$} &  41.5 \fs{$\pm 1.0$}\\  
         \cmidrule{2-7}
         ~ & 10 & Full & 66.2 \fs{$\pm 0.5$} & 64.8 \fs{$\pm 0.9$} & 65.4 \fs{$\pm 0.2$}&62.4 \fs{$\pm 0.9$}\\
         \cmidrule{2-7}
         ~& 50 & Full & 74.5 \fs{$\pm 0.1$} & 73.2 \fs{$\pm $0.3}& 72.8 \fs{$\pm 0.0$}& 73.2 \fs{$\pm 0.2$}\\
         \bottomrule[1.5pt]
    \end{tabular}
    \label{tab:app_std}
    \vspace{-1em}
\end{table}

\subsection{Comparison with Other Baselines} \label{sec:more_baselines}
\begin{table}[ht]
    \centering
    \vspace{-1em}
    \caption{Comparison with baselines. p in DropPath and Dropout denotes the keep rate and \textit{alpha} in MixUp and CutMix is the parameters $\alpha$ and $\beta$ in beta distribution, where $\alpha$ and $\beta$ are the same.}
    \scriptsize
    \tabcolsep=0.5em
    \begin{tabular}{c|c c c c c c}
    \toprule[1.5pt]
        Method & Baseline &\makecell{DropPath\\(p=0.5)} &\makecell{DropOut\\(p=0.5)} &\makecell{MixUp\\(alpha=0.5)} &\makecell{CutMix\\(alpha=0.5)} &\textbf{Ours}\\
        \midrule[0.5pt]
         Test Acc.& 55.6 & 46.2&44.3&57.6&56.9 & \textbf{66.2}\\
         \bottomrule[1.5pt]
    \end{tabular}
    \label{tab:more_baselines}
\end{table}
We further compare our method with some regularization methods on architectures, including DropOut \citep{srivastava2014dropout} and DropPath \citep{larsson2016fractalnet}, and data augmentation methods, including MixUp \citep{zhang2017mixup} and CutMix \citep{yun2019cutmix}. 
We use 3-layer CNN as the training network and ResNet18 as the test network. We use FRePo to generate distilled dataset and set $IPC$ to be $10$.
Our results are reported in Table \ref{tab:more_baselines}. We observe that DropPath and DropOut with a constant keep rate deteriorate the performance, compared with the DropPath variant proposed by us. 
In addition, MixUp and CutMix only contribute marginal performance improvement, compared with 2-fold augmentation in this work.
These results further demonstrate the effectiveness of our methods to train distilled datasets.

\subsection{Improve the Performance of Training on Limited Real Data} \label{sec:4.2}

In this section, we discuss the performance of our methods when training on a limited amount of real data and compare it with the case of the distilled dataset.
Our methods have shown effective on the distilled dataset, we expect them to improve the performance on limited real training data as well.
In this case, smaller models also tend to perform better than larger models because both can fit the training set perfectly but the latter suffers more from overfitting.

\begin{figure*}[!ht]
    \centering
    \subfigure[ResNet18 v.s. 3-layer CNN]{\includegraphics[width=0.23\textwidth]{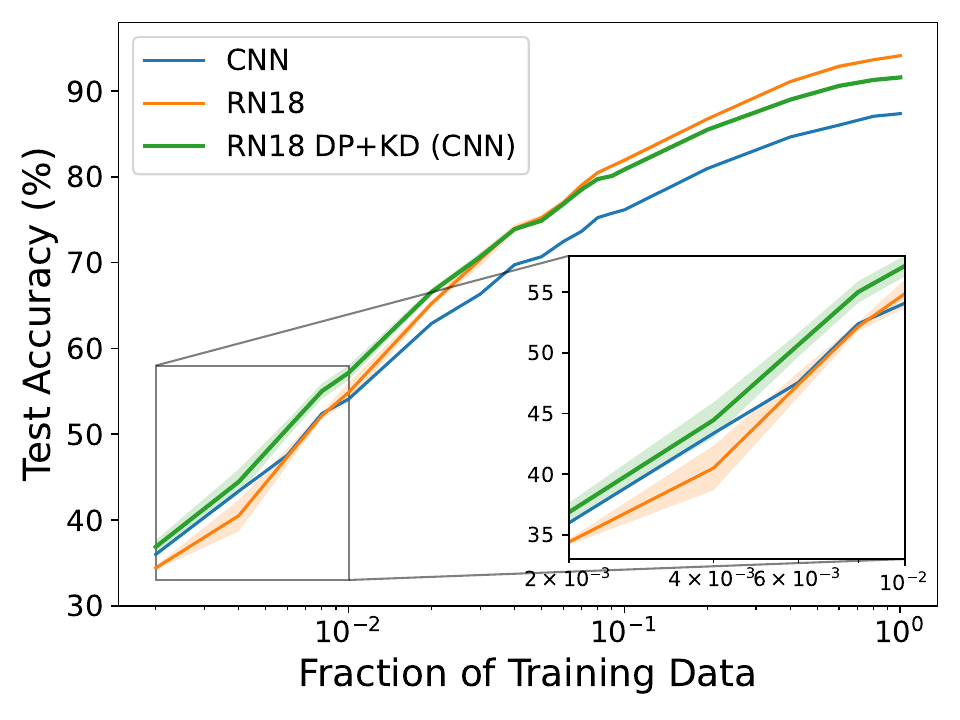}}
    \subfigure[ResNet50 v.s. 3-layer CNN]{\includegraphics[width=0.23\textwidth]{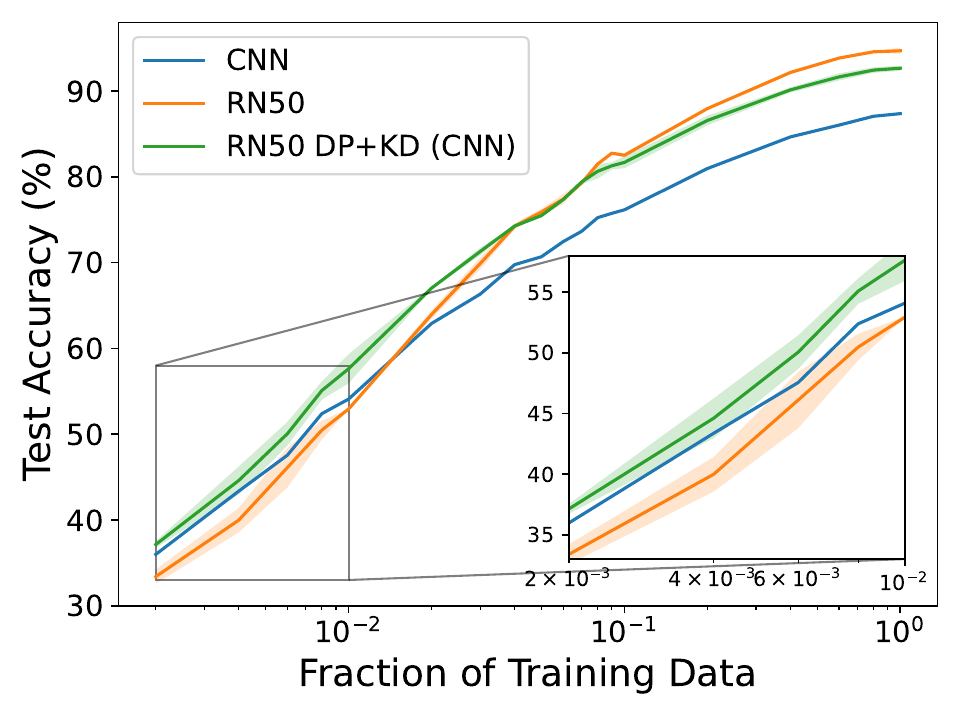}}
    \subfigure[VGG11 v.s. 3-layer CNN]{\includegraphics[width=0.23\textwidth]{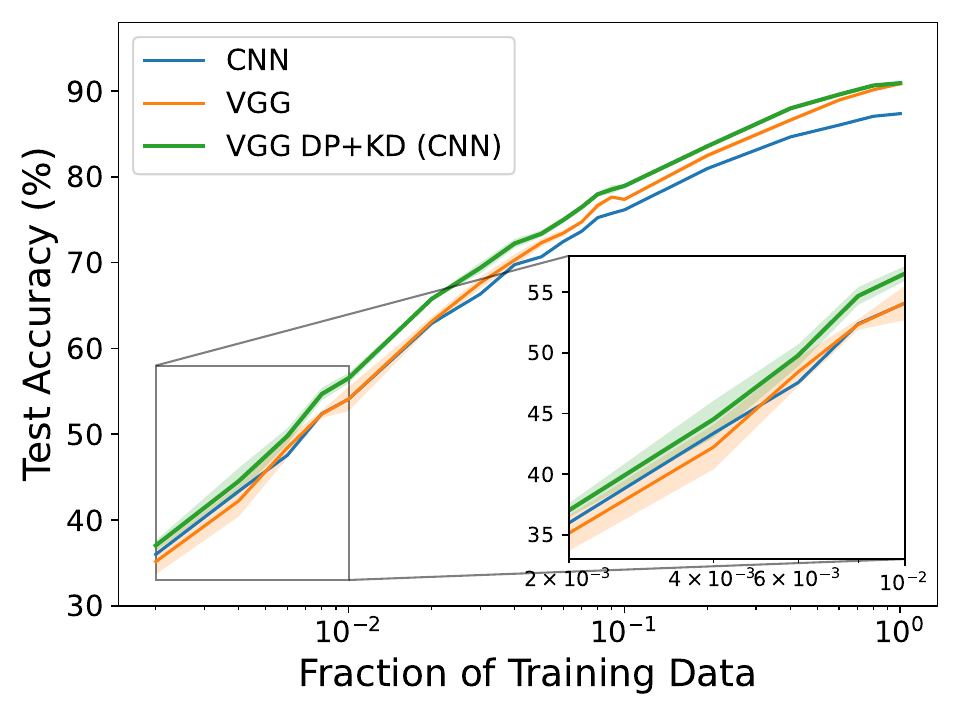}}
    \subfigure[ResNet50 v.s. ResNet18]{\includegraphics[width=0.23\textwidth]{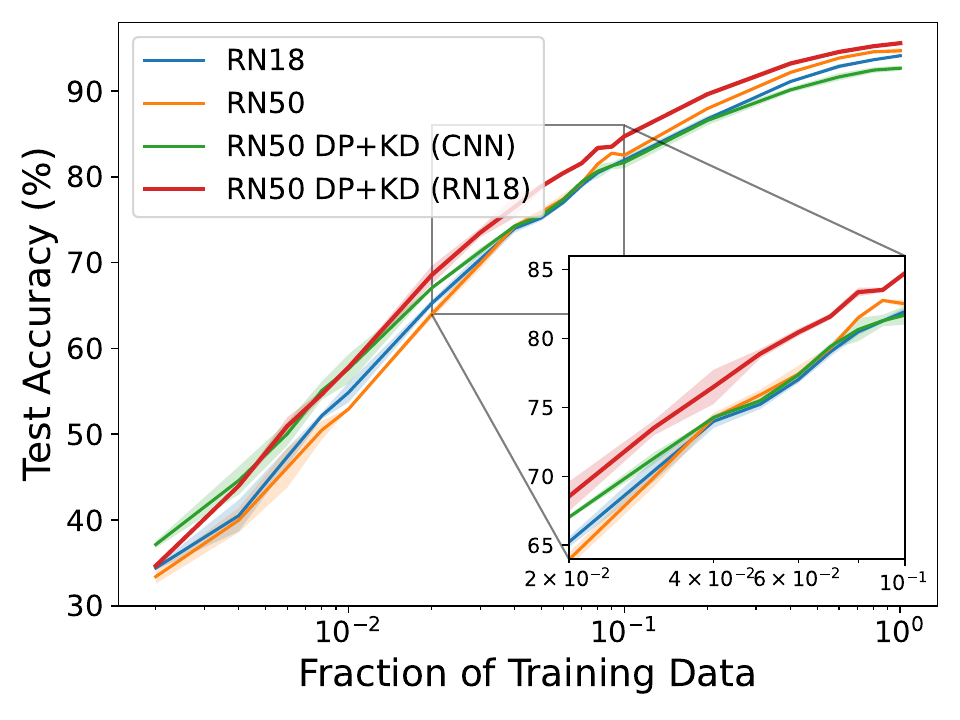}}
    \caption{\small Test accuracies obtained from training on different fractions of CIFAR10, the shadow indicates the standard deviation. We compare the test accuracies \textbf{(a)} between ResNet18 (RN18) and 3-layer CNN (CNN), \textbf{(b)} between ResNet50 (RN50) and CNN, \textbf{(c)} between VGG11 and 3-layer CNN (CNN), and \textbf{(d)} between ResNet50 (RN50) and ResNet18, respectively. The x-axis denotes the fraction of training data, 
    \textit{DP+KD} denotes that the network is trained with DropPath and knowledge distillation.
    The model enclosed in the brackets after KD represents the teacher model used.
    Note that we run the experiments three times with different random seeds.
    }
    \label{fig:4.2}    

\end{figure*}

As illustrated in Figure \ref{fig:4.2}, we train models on different fractions of CIFAR10 training set which are randomly sampled.
The 3-layer CNN still serves as the teacher model when we use knowledge distillation.
Since ResNet18 and ResNet50 exhibit the largest performance differences from the 3-layer CNN in the previous experiments, we only show the results of ResNet18 and ResNet50 here.
ResNet18 and ResNet50 significantly outperform 3-layer CNN with enough training data, but they show worse generalization performance than CNN when the fraction is lower than 0.02, i.e., $1000$ training instances.
Under our methods, the performances of both ResNet18 and ResNet50 surpass that of 3-layer CNN even when the fraction is as small as 0.002, i.e., 100 training instances.
However, the performance gain saturates or even declines when the fraction of training data exceeds $0.05$. This can be attributed to the suboptimal performance of the teacher model (blue line).
Nevertheless, Figure \ref{fig:4.2} (d) shows that when the current teacher does not contribute to performance gain anymore, a stronger teacher can further improve the performance.

Furthermore, we observe that the performance gap of training on limited real data is much smaller than that of training on distilled images.
For instance, when the fraction of training data is $0.002$, which is equivalent to IPC=10, the performance gap between 3-layer CNN and ResNet50 is $4.9\%$ when they are trained on real images. However, when we train them on distilled images of FRePo, the performance gap increases to $18.6\%$. As for the distilled images generated by MTT, the gap is even larger, which reaches $35.5\%$.
Meanwhile, training on a distilled dataset results in much better performance than training on real data of the same size, which makes it popular in downstream applications. 
Therefore, we focus on applying our method in the context of dataset distillation, in which the effectiveness of our method can be better revealed.

\subsection{Smoothing Effect Induced by Proposed Methods} \label{sec:smooth}
To corroborate the smoothing effect induced by our proposed methods, we analyze the Hessian spectrum of models trained with different ablations. It is known that the curvature in the neighborhood of model parameters is dominated by the top eigenvalues of the Hessian matrix $\nabla^2\mathcal{L}_{CE}(\theta)$, where $\mathcal{L}_{CE}(\theta)$ denotes the cross-entropy loss w.r.t model parameters $\theta$. In the implementation, we use the power iteration as in \citep{yao2018hessian, liu2020loss} to iteratively estimate the top 20 eigenvalues and the corresponding eigenvectors of the Hessian matrix.

\begin{figure*}[htbp]
    \centering
    \subfigure[Hessian Eigenvalues]{\includegraphics[width=0.31\textwidth]{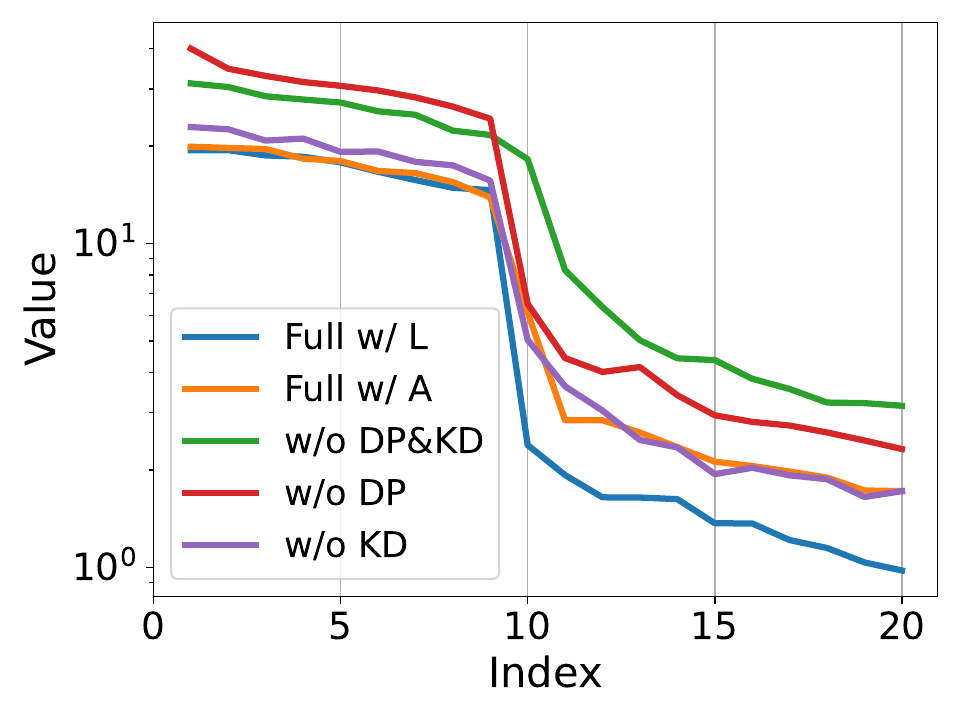}}
    \subfigure[Full w/ Lion]{\includegraphics[width=0.31\textwidth]{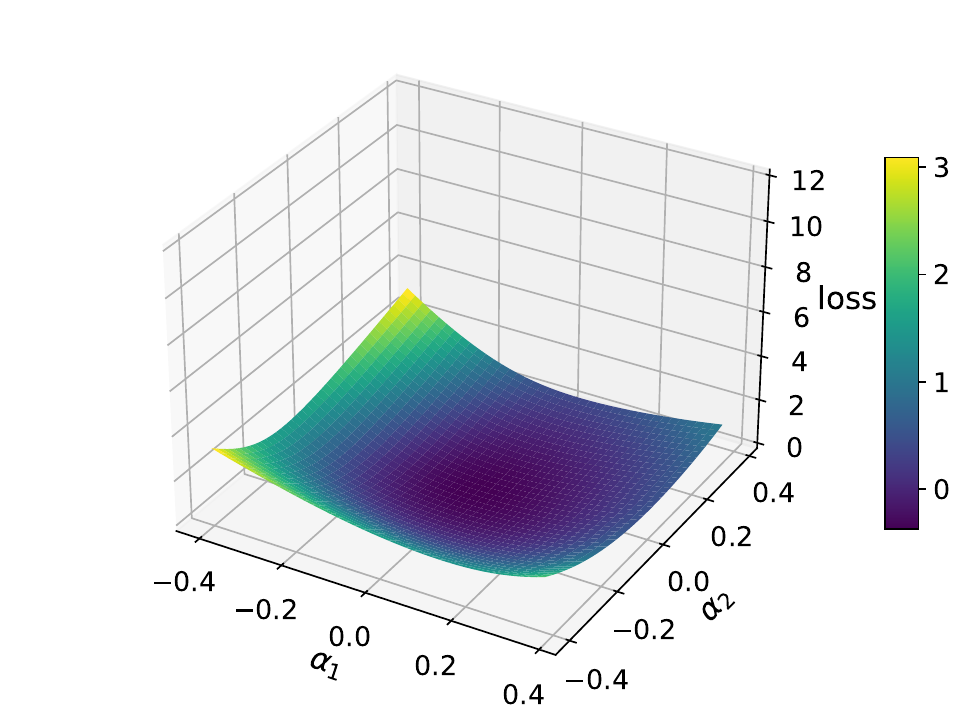}}
    \subfigure[Full w/ AdamW]{\includegraphics[width=0.31\textwidth]{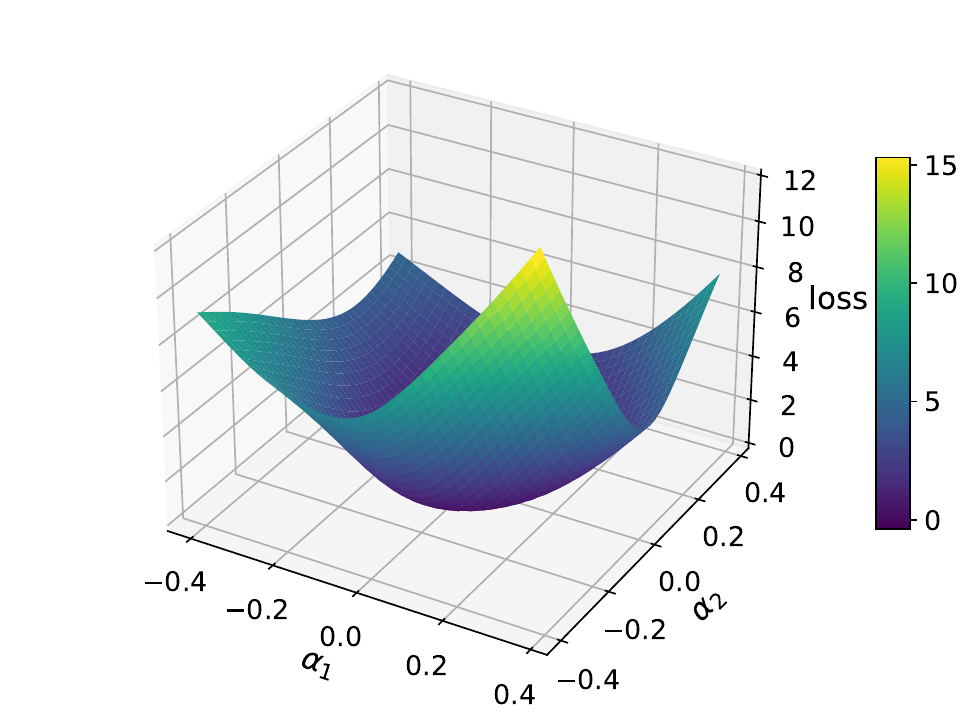}}\\
    \subfigure[w/o DP]{\includegraphics[width=0.31\textwidth]{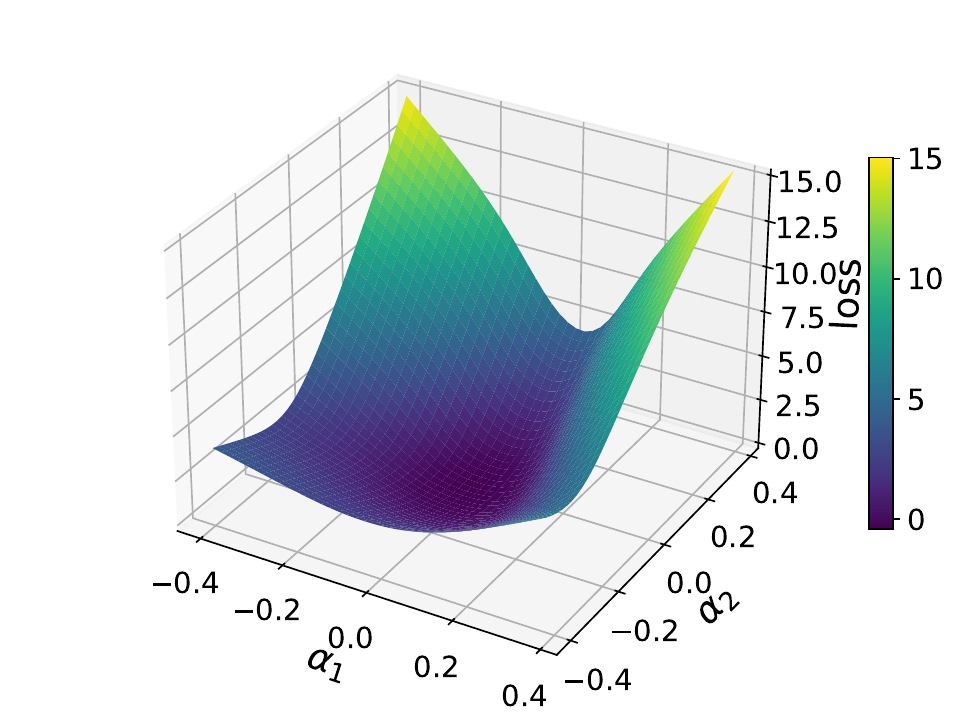}}
    \subfigure[w/o KD]{\includegraphics[width=0.31\textwidth]{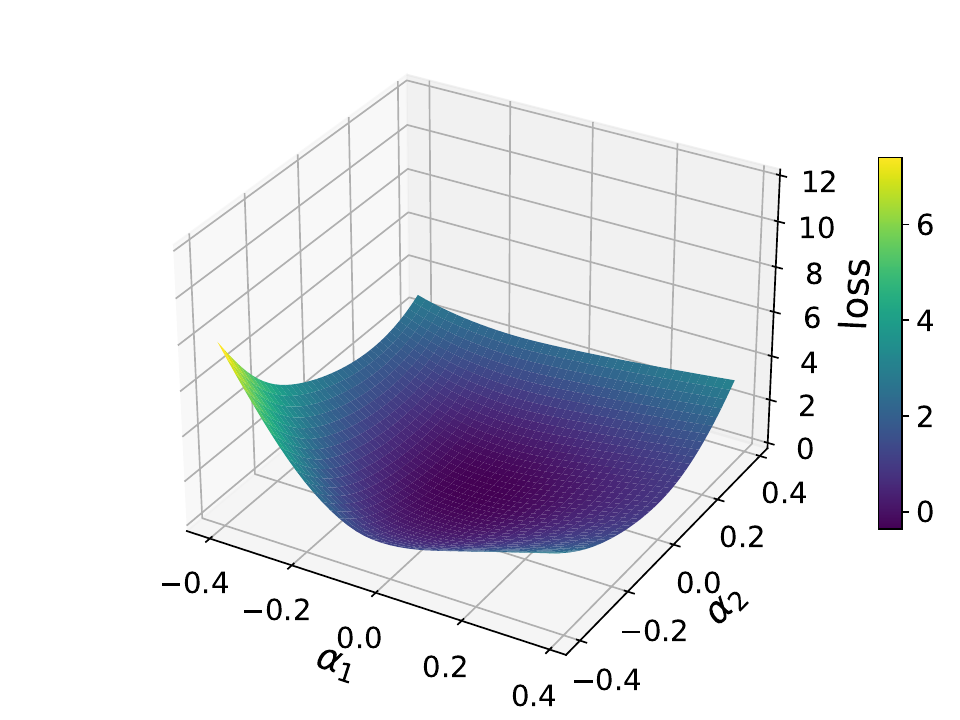}}
    \subfigure[w/o DP\&KD]{\includegraphics[width=0.31\textwidth]{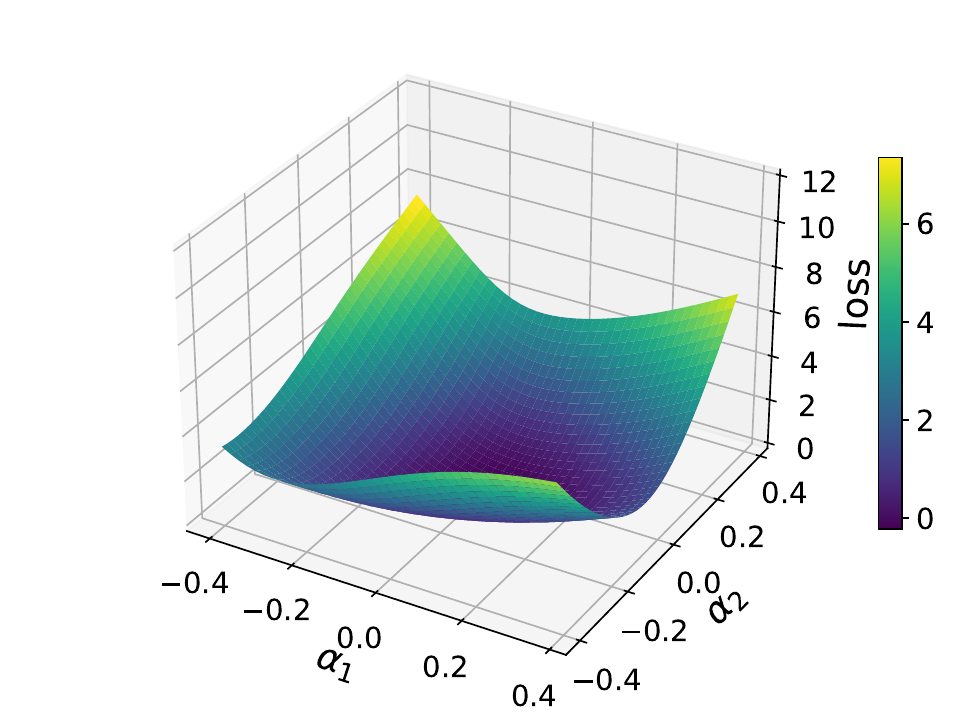}}
    
    \caption{\small Visualization of the smoothing effect induced by proposed methods. \textbf{(a)} Top 20 eigenvalues of Hessian matrix for ResNet18 trained with different settings, including full setting with Lion optimizer (Full w/ L), full setting with AdamW (Full w/ A), w/o DP, w/o KD and w/o DP\&KD. For w/o DP, w/o KD and w/o DP\&KD, Lion is adopted by default. \textbf{(b)-(f)} Loss landscape $\mathcal{L}_{CE}(\theta + \alpha_1\mathbf{v}_1 + \alpha_2\mathbf{v}_2)$ of ResNet18 around the minima found by models with different settings, where $\mathbf{v}_1$ and $\mathbf{v}_2$ are the eigenvectors corresponding to the top two eigenvalues of Hessian matrices, respectively. Note that the training data is 100 (IPC=10) distilled images of CIFAR10 by FRePo. ResNet18 is trained with DropPath and knowledge distillation. 3-layer CNN serves as the teacher model where knowledge distillation is adopted.}
    \label{fig:app_ablation}
    \vspace{-1em}
\end{figure*}

As shown in Figure \ref{fig:app_ablation} (a), the eigenvalues of the Hessian matrix for the ResNet18 trained with full setting and Lion optimizer are the lowest among the evaluated settings, which quantitatively indicates that the neighborhood of the minima found by our method has smaller curvature. Furthermore, Figure \ref{fig:app_ablation} (b)-(f) qualitatively shows that our method induces a smoother loss landscape. Notably, DropPath, forming an implicit ensemble of sub-networks, contributes the most to loss landscape smoothing. By contrast, knowledge distillation only has a marginal effect on smoothing.

\subsection{Ablation Studies}\label{sec:ablation}
We conduct extensive ablation studies here to validate the effectiveness of each component in our methods.
In this subsection, we focus on the case of using 3-layer CNN as the training network, ResNet18 as the test network, setting IPC to $10$ and generating the distilled dataset by FRePo.
Note that the baseline performance of 3-layer CNN trained on the distilled data is $63.0\%$, its performance improves to $64.7\%$ with better optimization and data augmentation.
\begin{figure*}[htb]
    \centering
    \subfigure[Minimum keep rate]{\includegraphics[width=0.19\textwidth]{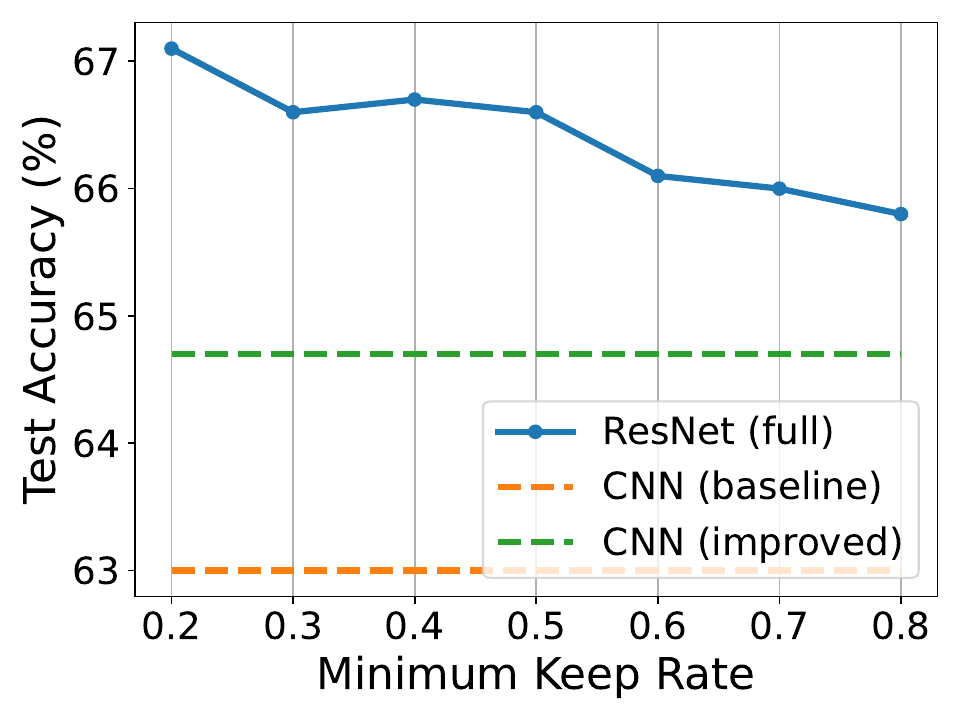}}
    \subfigure[Final keep rate]{\includegraphics[width=0.19\textwidth]{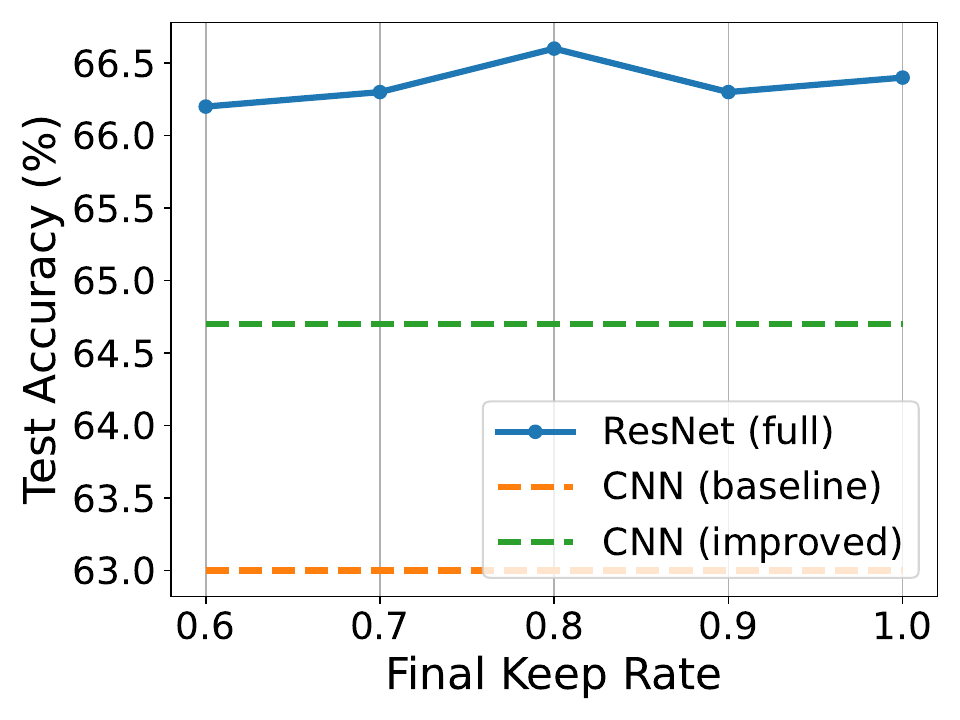}}
    \subfigure[Period of decay]{\includegraphics[width=0.19\textwidth]{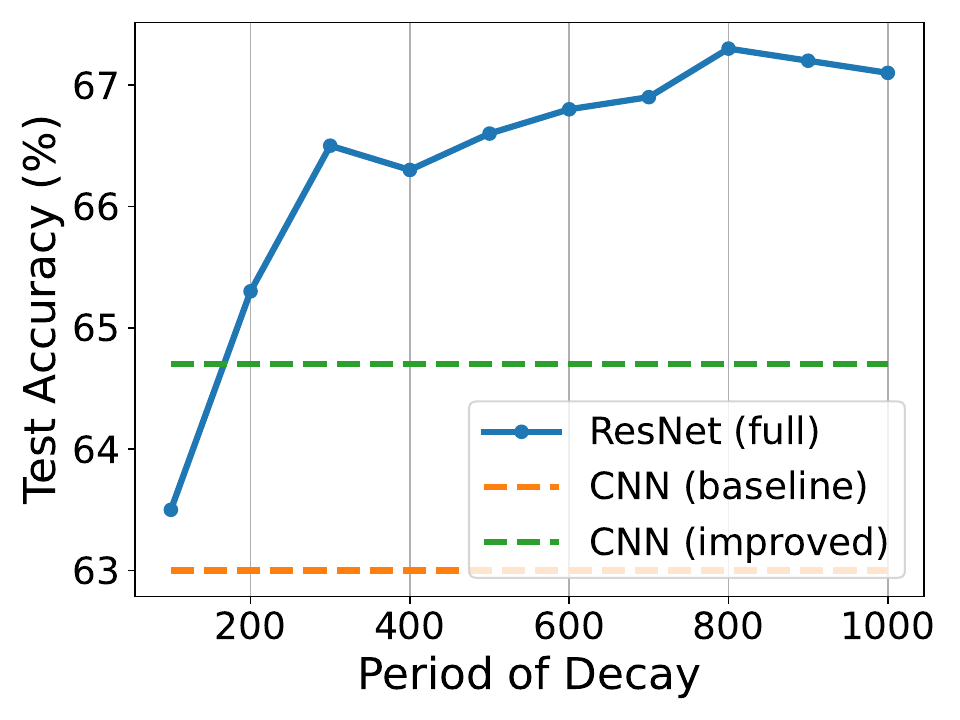}}
    \subfigure[KD weight]{\includegraphics[width=0.19\textwidth]{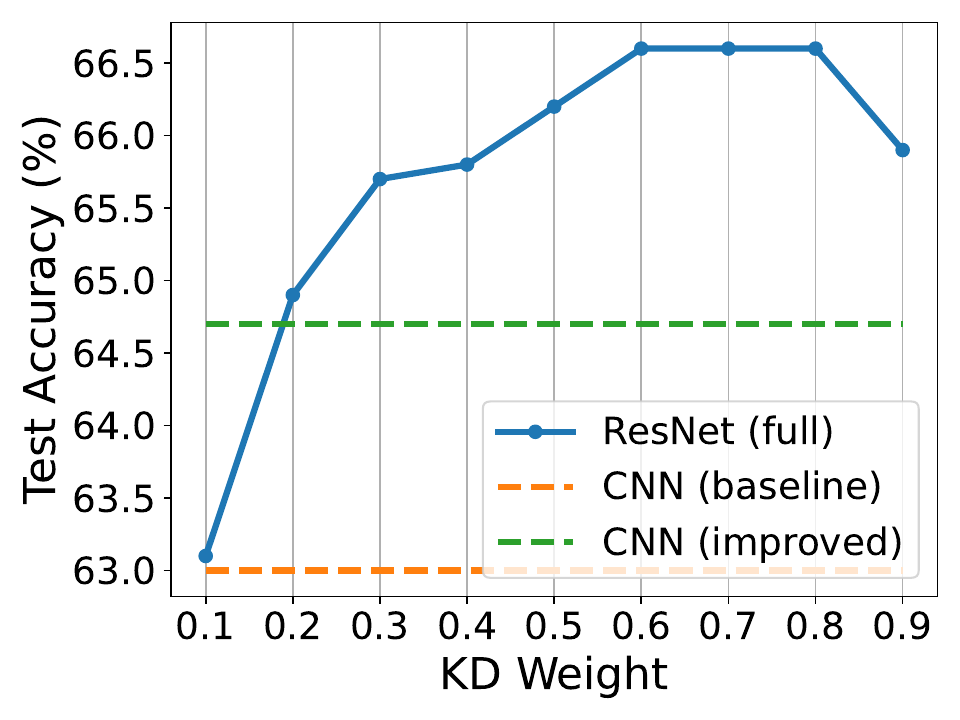}}
    \subfigure[KD temperature]{\includegraphics[width=0.19\textwidth]{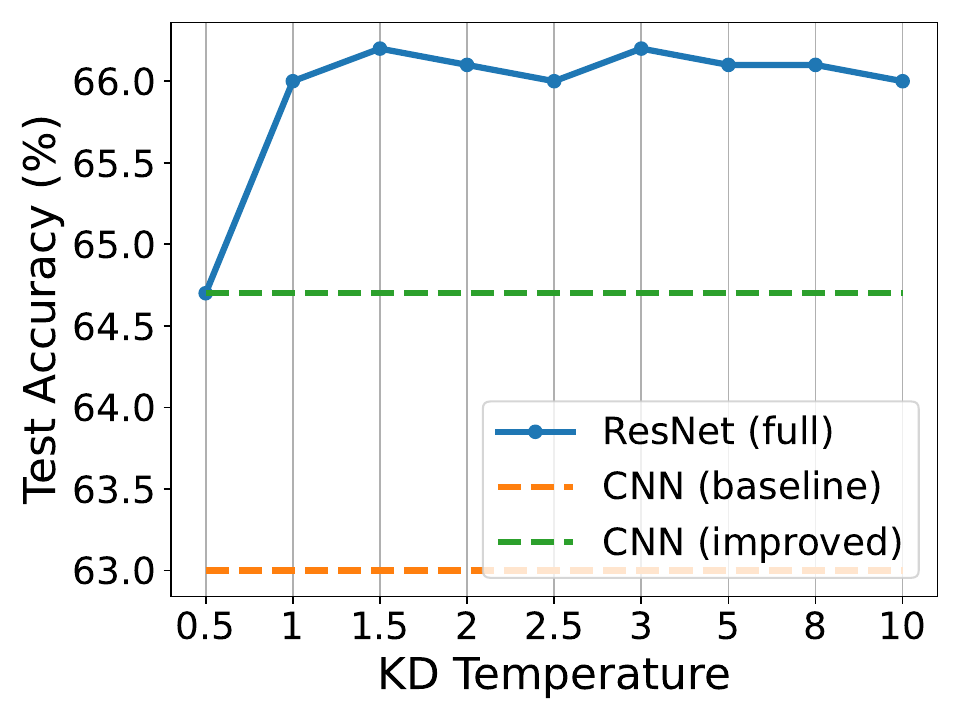}}
    \caption{\small Ablation studies on minimum keep rate, final keep rate, period of decay, weight and temperature of knowledge distillation (KD). \textbf{(a)} Test accuracies of different minimum keep rates. \textbf{(b)} Test accuracies of different keep rates at the final phase. \textbf{(c)} Test accuracies of different periods of decay. \textbf{(d)} Test accuracies of different KD weights. \textbf{(e)} Test accuracies of different KD temperatures.
    Regardless of the variation of hyperparameters, ResNet18 trained with our approach generally outperforms 3-layer CNN trained with baseline (orange dashed line) and that trained with better optimization and data augmentation (green dashed line).}
    \vspace{-1em}
    \label{fig:ablation}
\end{figure*}

\textbf{DropPath: }
We first try different minimum keep rates in the three-phase scheduler introduced in Section~\ref{sec:droppath}. As illustrated in Figure \ref{fig:ablation} (a) and (c), a lower minimum keep rate and a longer period of decay induce better performance, but both of them make the training longer. To balance performance and efficiency, we set the minimum keep rate and period of decay to $0.5$ and $500$, respectively. Figure \ref{fig:ablation} (b) shows that different final keep rates do not significantly affect the performance.
Moreover, we verify the effectiveness of the high keep rate in the final phase of training, and the improved shortcut connection (SC) introduced in Section~\ref{sec:droppath}. The results shown in Table \ref{tab:ablation_dp} indicate that both of them contribute to the performance.

\begin{table}[tb]
    \centering
    \caption{Ablation studies on the high keep rate in the final phase of training and improved shortcut connection (SC).
    }
    \small
    \label{tab:ablation_dp}
    \begin{tabular}{c c |c}
        \toprule[1.5pt]
        Final phase & Improved SC & Test Acc.\\
        \midrule[1pt]
        \redcross & \redcross &  65.2\\
        \greencheck & \redcross& 65.6\\
        \redcross & \greencheck &  65.9\\
        \greencheck & \greencheck & \textbf{66.6}\\
        \bottomrule[1.5pt]
    \end{tabular}
    \vspace{-1em}
\end{table}

\textbf{Knowledge Distillation: }
We also test different hyperparameters of knowledge distillation (KD). As illustrated in Figure \ref{fig:ablation} (d) and (e), when weight $\alpha$ and temperature $\tau$ are in the range of [0.5, 0.8] and [1, 10], respectively, the performance does not vary significantly. It indicates that our method is quite robust to different hyperparameter choices.
\begin{table}[t]
    \centering
    \caption{Ablation studies about optimization and data augmentation.
    If periodical learning rate (LR), Lion optimizer and stronger augmentation (Aug.) are not adopted, we replace them with cosine annealing learning rate, AdamW and 1-fold augmentation, respectively.
    }
    \label{tab:ablation_mt}
    \small
    \begin{tabular}{c c c|c}
        \toprule[1.5pt]
         Periodical LR & Lion & stronger Aug. & Test Acc.\\
         \midrule[1pt]
         \redcross & \redcross & \redcross &  61.6\\
         \greencheck & \redcross & \redcross &  61.9\\
         \greencheck & \greencheck & \redcross & 64.8\\
         \greencheck & \greencheck & \greencheck & \textbf{66.6}\\
         \bottomrule[1.5pt]
    \end{tabular}
    \vspace{-1em}
\end{table}

\textbf{Optimization and Data Augmentation:}
In Table~\ref{tab:ablation_mt}, we replace each of the optimization and data augmentation approaches with a baseline. The results indicate that each of these approaches improves performance. Among them, Lion optimizer contributes a performance improvement of $2.9\%$.

\textbf{Impact of Augmentation when IPC=$1$: }
It should be noted that the results of IPC=1 in Table~\ref{tab:cifar10} are obtained with 4-fold augmentation. For comparison, we also get the results with 2-fold augmentation (see in Table \ref{tab:app_impact_aug}). 
Compared with Table \ref{tab:cifar10}, the test accuracies of \textit{w/o DP\&KD} and \textit{w/o \& KD} in Table \ref{tab:app_impact_aug} are higher, but those of \textit{w/o DP} and \textit{Full} are lower. Especially for ResNet50, the performance of \textit{Full} increases by $7.7\%$ with 4-fold augmentation. This indicates that stronger augmentation is necessary when using knowledge distillation when there are extremely limited data, and when the architecture difference between the training and test networks is big. Moreover, we observe that the contribution of 4-fold augmentation is marginal under a larger IPC, so we adopt 4-fold augmentation only when IPC=1.

\begin{table}[ht]
\vspace{-1em}
    \centering
    \caption{Test accuracies of models trained on the distilled data of CIFAR10 (FRePo, IPC=1). However, 2-fold augmentation is adopted here. Except that, the other settings are the same as Table \ref{tab:cifar10}.}
    \tabcolsep=0.5em
    \scriptsize
    \begin{tabular}{c|c|c|c c c c}
        \toprule[1.5pt]
         IPC & Methods & \makecell{3-layer\\CNN} & ResNet18 & AlexNet & VGG11 & ResNet50\\
         \midrule[1pt]
          \multirow{5}{*}{1} & Baseline & 44.3 & 34.4 \fs{(\red{-9.9})} & 41.8 \fs{(\red{-2.5})} & 44.0 \fs{(\red{-0.3})} & 25.9 \fs{(\red{-18.4})}\\
          ~ & w/o DP\&KD & \textbf{44.8} \fs{(\blue{+0.5})} & 41.2 \fs{(\red{-3.1})} & 45.4 \fs{(\blue{+1.1})} & 45.9 \fs{(\blue{+1.6})} & 32.8 \fs{(\red{-11.5})}\\
           ~ & w/o DP & - & 41.0 \fs{(\red{-3.3})} & 44.5 \fs{(\blue{+0.2})} & \textbf{47.0} \fs{(\blue{+2.7})} & 30.0 \fs{(\red{-14.3})}\\
         ~ & w/o KD & - & 39.4 \fs{(\red{-4.9})} & 47.1 \fs{(\blue{+2.8})} & 38.9 \fs{(\red{-5.4})} & 31.0 \fs{(\red{-13.3})}\\
         ~ & \cellcolor{gray!40} \textbf{Full} & \cellcolor{gray!40}- & \cellcolor{gray!40} \textbf{45.5} \fs{(\blue{+1.2})} & \cellcolor{gray!40}\textbf{47.8} \fs{(\blue{+3.5})} & \cellcolor{gray!40}46.7 \fs{(\blue{+2.4})} & \cellcolor{gray!40} \textbf{33.8} \fs{(\red{-10.5})}\\  
         \bottomrule[1.5pt]
    \end{tabular}
    \label{tab:app_impact_aug}
    \vspace{-1em}
\end{table}

\section{Conclusion}
This paper studies architecture overfitting when we train models on distilled datasets.
To mitigate this issue, we propose a series of approaches based on the intuition that the large model can act as an implicit ensemble of small models. These methods also exhibit a smoothing effect from different aspects. Our methods are efficient and generic, and can improve the performance when training on a small real dataset directly. We believe that our work can help extend the utility of distilled datasets in more real-world scenarios.
Recognizing that this work only mitigates architecture overfitting in the evaluation stage, our future work will focus on developing a more generalizable dataset distillation algorithm to address this issue in essence.
\section*{Acknowledgments}

This work is supported by the internal funds of City University of Hong Kong (No. 9610614 and No. 9229130). It is also supported by the NSFC project (No. 62306250). We also thank Shuqi Liu for her support in experiments.
\bibliographystyle{IEEEtranN}
\bibliography{references}
\appendix

\subsection{Implementation Details} \label{sec:app_imple}
\textbf{Datasets:} The training sets in the experiments are the distilled datasets of CIFAR10, CIFAR100 \citep{krizhevsky2009learning} and Tiny-ImageNet \citep{deng2009imagenet}, but the test sets are their respective original test sets. To better validate the effectiveness of our method, we use the distilled images synthesized by different dataset distillation algorithms, e.g., neural Feature Regression with Pooling (FRePo) \citep{zhou2022dataset} and Matching Training Trajectories (MTT) \citep{cazenavette2022dataset}. In addition, we evaluate the performance of our method in different numbers of instances per class (IPC), e.g., 1, 10 and 50. Note that MTT does not provide the final trainable learning rates in the released checkpoints, we adopt the reported initial learning rates in our baselines.

\textbf{Models:} The networks used to synthesize the distilled images (training networks) in FRePo and MTT are 3-layer CNN. Consistent with the hyperparameters reported in the paper, the output channels of hidden layers of the network used in FRePo are 128, 256 and 512, respectively. However, in MTT, all the output channels of hidden layers are set to 128. ResNet18, ResNet50 \citep{he2016deep}, AlexNet \citep{NIPS2012_c399862d} and VGG11 \citep{simonyan2014very} are adopted in the evaluation, they are thereby called test networks. The hyperparameters of networks are the same as those set in \citep{cazenavette2022dataset}. Note that when training networks on distilled images of FRePo and MTT, batch and instance normalization layers are adopted in networks following the settings of \citep{zhou2022dataset, cazenavette2022dataset}, respectively.

\textbf{DropPath:} As shown in Algorithm \ref{alg:droppath}, the decaying factor of keep rate $\gamma=0.1$, minimum keep rate $p_{min}=0.5$, final keep rate $p_{final}=0.8$, period of decay $T=500$, warmup period $W=50$, stabilization epoch $S=(1+p_{min}/\gamma)\times T=3000$. The total epochs $N$ is set to $S+2\times T = 4000$. In the improved shortcut, the pooling area depends on the stride of $1\times1$ convolutional layer in the original one. e.g., if the stride of $1\times1$ convolutional layer in the original shortcut is $2$, we use a $2\times2$ max pooling.

\textbf{Knowledge distillation:} As shown in Eq. \ref{eq:kd}, the temperature factor $\tau=1.5$, and the weight factor $\alpha=0.5$. If not specifically indicated, the default teacher model is the 3-layer CNN. Note that the teacher model is trained on the same data set as the student model.

\textbf{Periodical learning rate:} In Eq. \ref{eq: lr}, the maximum learning rate $lr_{max}=5\times10^{-5}$, the base decaying factor for learning rate $\lambda=0.8$. The period of the cosine function $T_{max}$ and the number of warmup epochs $T_{warm}$ are $1000$ and $50$, respectively.
    
\textbf{Optimizer:} Lion \citep{chen2023symbolic} is adopted in our method, where weight decay $\lambda_{wd}=5\times10^{-3}$, coefficient $\beta_1=0.95$, and $\beta_2=0.98$.

\textbf{Augmentation:} There are color jittering, cropping, cutout, flipping, scaling, and rotating in the augmentation pool, we sample more operations instead of just one as in \citep{cazenavette2022dataset}.

\textbf{Training:} For CIFAR10, the batch sizes for different IPCs are 10 (IPC=1), 100 (IPC=10) and 128 (IPC=50), respectively. For CIFAR100, the batch sizes are 100 (IPC=1), 256 (IPC=10) and 256 (IPC=50), respectively. Cross-entropy is adopted as the loss function in our experiments. Since the labels of images are learnable in FRePo, we divide them with a temperature factor $t=0.3$ for CIFAR10, $0.04$ for CIFAR100, and $0.02$ for Tiny-ImageNet, respectively.

\subsection{Supplementary Figures of Methods} \label{sec:app_droppath}
\begin{figure}[ht]
\vspace{-1em}
    \centering
    \subfigure[]{\includegraphics[width=0.24\textwidth]{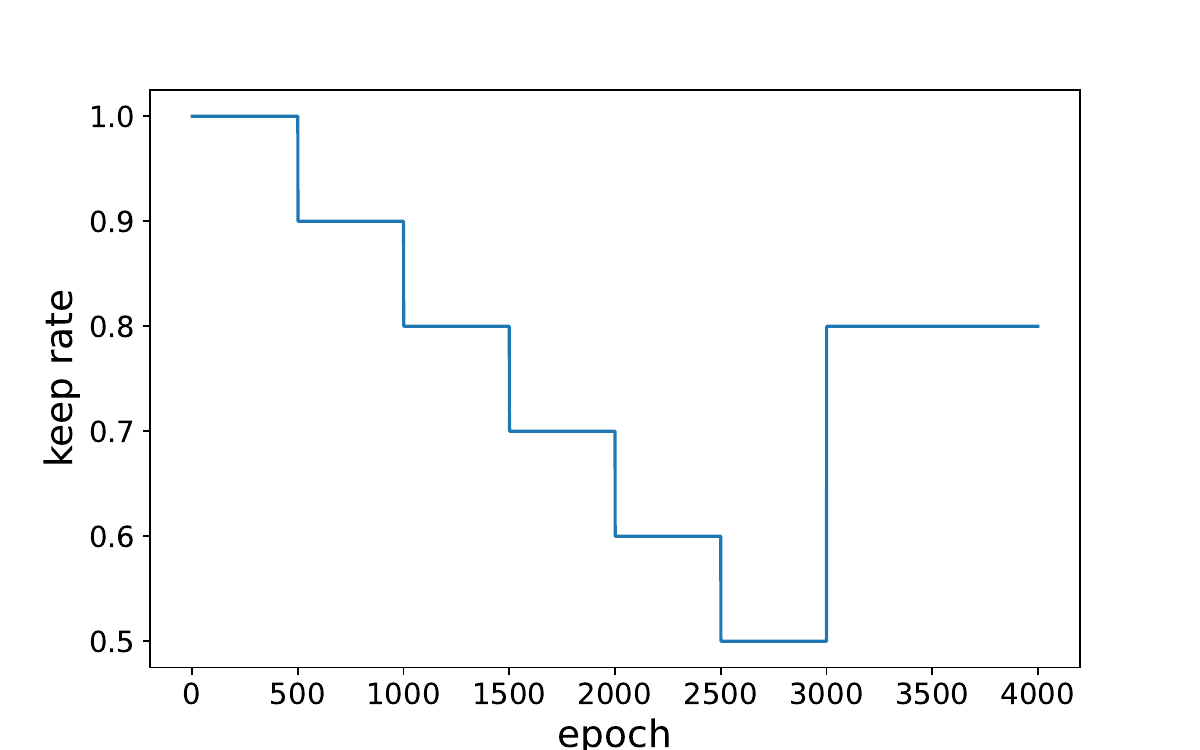}}
    \subfigure[]{\includegraphics[width=0.24\textwidth]{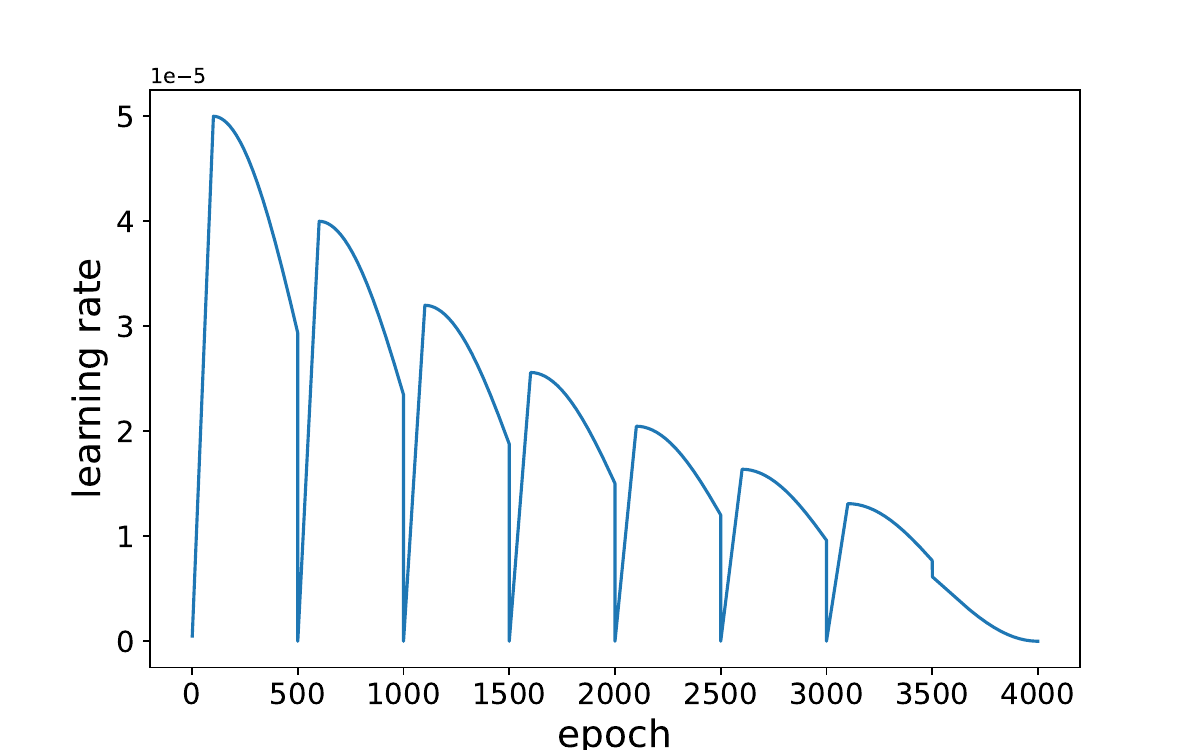}}
    \caption{Supplementary figures. \textbf{(a)} Scheduler of three-phase keep rate. \textbf{(b)} Curve of periodical learning rate.}
    \label{fig:kr_lr}
    \vspace{-1em}
\end{figure}

The corresponding curve of the dynamic keep rate is shown in Figure \ref{fig:kr_lr} (a). Mathematically, the dynamic keep rate $p$ is formulated as 

\begin{equation} \label{eq: keep_rate}
    p = 
    \begin{cases}
         \mathtt{max}(p_{\mathrm{min}}, 1-\gamma\cdot \mathtt{ceil}((i - W)/T)) , ~~~~~~~\text{if $i<S$}, \\
        p_{\mathrm{final}},~~~~~~~~~~~~~~~~~~~~~~~~~~~~~~~~~~~~~~~~~~~ \text{otherwise}.
    \end{cases}
\end{equation}
where $\gamma\in[0,1]$ is a decaying factor. $i$, $W$, $T$ and $S$ denote the current epoch, warmup period, decay period and stabilization epoch, respectively. Unless specified, we set $\gamma = 0.1$, $p_{\mathrm{min}} = 0.5$, $p_{\mathrm{final}} = 0.8$, $T = 500$, $W = 500$, $S = 3000$ in the experiments.

Figure \ref{fig:kr_lr} (b) shows how the periodical learning rate changes.
The learning rate $lr$ at epoch $i$ is defined as 
\begin{equation} \label{eq: lr}
    \mathrm{lr}_i = 
    \begin{cases}
         \lambda_i \cdot \frac{\mathrm{mod}(i, t)}{T_{\mathrm{warm}}} \cdot \mathrm{lr}_{\mathrm{max}} , ~~~~~~~~~~~~\text{if $\mathrm{mod}(i, t)\leq T_{\mathrm{warm}}$}, \\
        0.5\lambda_i(1+\mathrm{cos}(\pi\frac{\mathrm{mod}(i, t) - T_{\mathrm{warm}}}{T_{\mathrm{max}}-T_{\mathrm{warm}}}))\cdot \mathrm{lr}_{\mathrm{max}},~ \text{otherwise}.
    \end{cases}
\end{equation}
where $T$ is the decay period of the keep rate $p$ of DropPath, $S$ is the stabilization epoch. $t=T$ when $i<S$, otherwise $t=S$. $\lambda_i = \lambda^{\lfloor \mathtt{min}(i, S)/T \rfloor}$ where $\lambda$ is a base decaying factor, and $\lfloor\cdot\rfloor$ denotes the floor function. $\mathrm{lr}_{\mathrm{max}}$ denotes the maximum learning rate, $\mathrm{mod}(x,y)$ denotes the remainder of $x/y$. The maximum iterations of the cosine annealing function and the number of warmup epochs are denoted by $T_{\mathrm{max}}$ and $T_{\mathrm{warm}}$, respectively. In implementation, the maximum learning rate $lr_{max}=5\times10^{-5}$, the base decaying factor for learning rate $\lambda=0.8$. The period of the cosine function $T_{max}$ and the number of warmup epochs $T_{warm}$ are $1000$ and $50$, respectively.

\subsection{Effect of scaling factor $1/p$ in DropPath} \label{sec:ap_scale_factor}
For multi-branch networks, Eq. \ref{eq:dp} shows that $\mathtt{DropPath}(\mathbf{x}) = \frac{m}{p}\cdot \vx,\quad m=\mathtt{Bernoulli}(p)$, where $p \in [0, 1]$ denotes the keep rate, $m = \mathtt{Bernoulli}(p) \in \{0,1\}$ outputs 1 with probability $p$ and 0 with probability $1-p$. 
We consider the module output $\vy = \mathtt{DropPath}(\vx)$ in the training phase, then the expectation of $\vy$ given $\vx$ is $\mathbb{E}(\vy) = p \cdot \frac{1}{p} \cdot \vx + (1 - p) \cdot \frac{0}{p} \cdot \vx = \vx$.
In the test phase, DropPath is disabled, so the module output is simply $\vx$ and consistent with the expectation in the training phase.
If there is no scaling factor $1 / p$ in Eq. \ref{eq:dp} and $p < 1$, the expectation of the module outputs in the training and test phases will be different, which leads to performance degradation.

However, virtual shortcut connection is adopted in single-branch networks, so the implementation of DropPath is different from that in multi-branch networks. In single-branch networks, the DropPath affects both main and shortcut paths. Therefore, given an input $\vx$ and the output of the main path $\vx'$ , if scaling factor is not considered here, the formulation of DropPath can be rewritten as $\mathtt{DropPath}(\mathbf{x}) = m\cdot \vx' + (1-m)\cdot \vx,\quad m=\mathtt{Bernoulli}(p)$. Assume that the expectations of $\vx'$ and $\vx$ are the same, the expectation of module output $\vy$ given $\vx$ is $\mathbb{E}(\vy) =  m\cdot \mathbb{E}(\vx') + (1-m)\cdot \vx = m\cdot \vx + (1-m)\cdot \vx = \vx$, which is not changed by $m$. As a result, the scaling factor is not necessary here.


\subsection{Results on Tiny-ImageNet} \label{sec:app_cifar100}
The results on Tiny-ImageNet are reported in \ref{tab:timage}, respectively. The observations on CIFAR10 and CIFAR100 are analogous to those on and Tiny-ImageNet, which indicates that our method is effective on different datasets. 

\begin{table*}[tb]
    \centering
    \caption{Test accuracies of models trained on the distilled data of \textbf{Tiny-ImageNet} \citep{deng2009imagenet} with different IPCs. 3-layer CNN is the architecture used for data distillation and is the teacher model of KD. Note that for DATM (IPC=50), the teacher model of ResNet50 is ResNet18 w/o DP\&KD.}
    \vspace{-1em}
    \small
    \begin{tabular}{c|c|c|c|c c c c}
        \toprule[1.5pt]
         DD & IPC & Methods & CNN & ResNet18 & AlexNet & VGG11 & ResNet50\\
         \midrule[1pt]
          \multirow{10}{*}{\rotatebox{90}{FRePo \citep{zhou2022dataset}}}& \multirow{5}{*}{1} & Baseline & 16.6 & 15.6 \scr{(\red{-1.0})} & 16.5 \scr{(\red{-0.1})} &  16.6 \scr{(\blue{+0.0})} & 13.4 \scr{(\red{-3.2})}\\
         ~ & ~ & w/o DP\&KD &  \textbf{17.7} \scr{(\blue{+1.1})} & 12.3 \scr{(\red{-4.3})} & 13.7 \scr{(\red{-2.9})} & 14.1 \scr{(\red{-2.5})} & 12.8 \scr{(\red{-3.8})}\\
          ~ & ~ & w/o DP & - & 15.8 \scr{(\red{-0.8})} & 16.6 \scr{(\blue{+0.0})} & 16.4 \scr{(\red{-0.2})} & 16.6 \scr{(\blue{+0.0})}\\
         ~ & ~ & w/o KD & - & 12.5 \scr{(\red{-4.1})} & 14.9 \scr{(\red{-1.7})} & 13.6 \scr{(\red{-3.0})} & 11.9 \scr{(\red{-4.7})}\\
         ~ & ~ & \cellcolor{gray!40} \textbf{Full} & \cellcolor{gray!40}- & \cellcolor{gray!40} \textbf{18.9} \scr{(\blue{+2.3})} & \cellcolor{gray!40}\textbf{18.5} \scr{(\blue{+1.9})} & \cellcolor{gray!40}\textbf{18.3} \scr{(\blue{1.7})} & \cellcolor{gray!40} \textbf{19.1} \scr{(\blue{+2.5})}\\ 
         \cmidrule{2-8}
         ~ & \multirow{5}{*}{10} & Baseline & \textbf{24.9} & 24.2 \scr{(\red{-0.7})} & 24.8 \scr{(\red{-0.1})}& 25.2 \scr{(\blue{+0.3})}& 24.9 \scr{(\blue{+0.0})}\\
          ~ & ~ & w/o DP\&KD & 23.0 \scr{(\red{-1.9})} & 21.7 \scr{(\red{-3.2})} & 23.8 \scr{(\red{-1.1})} & 24.2 \scr{(\red{-0.7})} & 23.1 \scr{(\red{-1.8})}\\
          ~ & ~ & w/o DP & - & 25.4 \scr{(\blue{+0.5})} & \textbf{25.2} \scr{(\blue{+0.3})} & 26.4 \scr{(\blue{+1.5})} & 26.9 \scr{(\blue{+2.0})}\\
          ~ & ~ & w/o KD & - & 21.5 \scr{(\red{-3.4})} & 22.4 \scr{(\red{-2.5})}& 24.0 \scr{(\red{-0.9})}& 21.6 \scr{(\red{-3.3})}\\
         ~ & ~ & \cellcolor{gray!40}\textbf{Full} & \cellcolor{gray!40}- & \cellcolor{gray!40}\textbf{26.8} \scr{(\blue{+1.9})} & \cellcolor{gray!40}24.9 \scr{(\blue{+0.0})} & \cellcolor{gray!40}\textbf{26.6} \scr{(\blue{+1.7})}& \cellcolor{gray!40}\textbf{27.3} \scr{(\blue{+2.4})}\\ 
         \midrule[1pt]
         \multirow{10}{*}{\rotatebox{90}{MTT \citep{cazenavette2022dataset}}}& \multirow{5}{*}{1} & Baseline & 8.8 & 6.2 \scr{(\red{-2.6})} &  6.7 \scr{(\red{-2.1})} & 7.3 \scr{(\red{-1.5})}& 2.7 \scr{(\red{-6.1})}\\
         ~ & ~ & w/o DP\&KD & \textbf{9.6} \scr{(\blue{+0.8})} & 6.1 \scr{(\red{-2.7})} & 8.4 \scr{(\red{-0.4})} & 7.2 \scr{(\red{-1.6})} & 3.2 \scr{(\red{-5.6})}\\
         ~ & ~ & w/o DP & - & 6.5 \scr{(\red{-2.3})} & 9.1 \scr{(\blue{+0.3})}& 7.9 \scr{(\red{-0.9})}& 3.6 \scr{(\red{-5.2})}\\
         ~ & ~ & w/o KD & - &  6.7 \scr{(\red{-2.1})} & 8.1 \scr{(\red{-0.7})} & 6.8 \scr{(\red{-2.0})} & 4.0 \scr{(\red{-4.8})}\\
         ~ & ~ &\cellcolor{gray!40} \textbf{Full} & \cellcolor{gray!40}- & \cellcolor{gray!40}\textbf{8.1} \scr{(\red{-0.7})} & \cellcolor{gray!40}\textbf{9.2} \scr{(\blue{+0.4})}& \cellcolor{gray!40}\textbf{8.2} \scr{(\red{-0.6})} & \cellcolor{gray!40}\textbf{8.2} \scr{(\red{-0.6})}\\
         \cmidrule{2-8}
         ~& \multirow{5}{*}{10} & Baseline & 19.3 & 17.2 \scr{(\red{-2.1})} & 14.3 \scr{(\red{-5.0})} & 15.1 \scr{(\red{-4.2})} & 11.2 \scr{(\red{-8.1})}\\
         ~ & ~ & w/o DP\&KD & \textbf{20.1}\scr{(\blue{+0.8})} & 16.6 \scr{(\red{-2.7})} & 18.7 \scr{(\red{-0.6})} & 16.2 \scr{(\red{-3.1})} & 15.2 \scr{(\red{-4.1})}\\
         ~ & ~ & w/o DP & - & 17.3 \scr{(\red{-2.0})} & 21.2 \scr{(\blue{+1.9})} & 19.9 \scr{(\blue{+0.6})} & 18.7 \scr{(\red{-0.6})}\\
         ~ & ~ & w/o KD & - & 19.0 \scr{(\red{-0.3})} & 17.7 \scr{(\red{-1.6})} & 15.2 \scr{(\red{-4.1})} & 17.7 \scr{(\red{-1.6})}\\
         ~ & ~ & \cellcolor{gray!40}\textbf{Full} & \cellcolor{gray!40}- & \cellcolor{gray!40}\textbf{22.6} \scr{(\blue{+3.3})} & \cellcolor{gray!40}\textbf{21.6} \scr{(\blue{+2.3})}& \cellcolor{gray!40}\textbf{20.5} \scr{(\blue{+1.2})}& \cellcolor{gray!40}\textbf{21.8} \scr{(\blue{+2.5})}\\  
         \midrule[1pt]
        \multirow{10}{*}{\rotatebox{90}{DATM \cite{guotowards}}}& \multirow{5}{*}{10} & Baseline & 16.0 & 13.8 \scr{(\red{-2.2})} & 16.0 \scr{(\blue{+0.0})} & 14.7 \scr{(\red{-1.3})} & 10.6 \scr{(\red{-5.4})} \\
        ~ & ~ & w/o DP\&KD & 18.4 \scr{(\blue{+2.4})} & 13.7 \scr{(\red{-2.3})} & 17.2 \scr{(\blue{+1.2})} & 16.4 \scr{(\blue{+0.4})} & 13.7 \scr{(\red{-2.3})} \\
        ~ & ~ & w/o DP & - & 13.8 \scr{(\red{-2.2})} & 17.8 \scr{(\blue{+1.8})} & 18.2 \scr{(\blue{+2.2})} & 15.3 \scr{(\red{-0.7})} \\
        ~ & ~ & w/o KD & - & 16.3 \scr{(\blue{+0.3})} & 15.9 \scr{(\red{-0.1})} & 17.6 \scr{(\blue{+1.6})} & 12.7 \scr{(\red{-3.3})} \\
        ~ & ~ & \cellcolor{gray!40} \textbf{Full} & \cellcolor{gray!40}- & \cellcolor{gray!40} \textbf{17.3} \scr{(\blue{+1.3})} & \cellcolor{gray!40}\textbf{17.6} \scr{(\blue{+1.6})} & \cellcolor{gray!40}\textbf{18.2} \scr{(\blue{+2.2})} & \cellcolor{gray!40} \textbf{15.9} \scr{(\red{-0.1})} \\ 
        \cmidrule{2-8}
        ~ & \multirow{5}{*}{50} & Baseline & 23.5 & 27.7 \scr{(\blue{+4.2})} & 28.4 \scr{(\blue{+4.9})}& 24.6 \scr{(\blue{+1.1})}& 21.6 \scr{(\red{-1.9})} \\
        ~ & ~ & w/o DP\&KD & 24.8 \scr{(\blue{+1.3})}& 29.2 \scr{(\blue{+5.7})} & 29.6 \scr{(\blue{+6.1})} & 25.3 \scr{(\blue{+1.8})} & 22.8 \scr{(\red{-0.7})} \\
        ~ & ~ & w/o DP & - & 28.2 \scr{(\blue{+4.7})} & 29.8 \scr{(\blue{+6.3})} & 27.7 \scr{(\blue{+4.2})} & 26.1 \scr{(\blue{+2.6})} \\
        ~ & ~ & w/o KD & - & 28.5 \scr{(\blue{+5.0})} & 28.2 \scr{(\blue{+4.7})} & 26.5 \scr{(\blue{+3.0})} & 23.2 \scr{(\red{-0.2})} \\
        ~ & ~ & \cellcolor{gray!40}\textbf{Full} & \cellcolor{gray!40}- & \cellcolor{gray!40}\textbf{29.9} \scr{(\blue{+6.4})} & \cellcolor{gray!40}\textbf{30.1} \scr{(\blue{+6.6})} & \cellcolor{gray!40}\textbf{28.3} \scr{(\blue{+4.8})} & \cellcolor{gray!40}\textbf{26.9} \scr{(\blue{+3.4})} \\ 
         \bottomrule[1.5pt]
    \end{tabular}
    \label{tab:timage}
\end{table*}

\end{document}